\documentclass[11pt]{article}

\usepackage[preprint]{acl}

\usepackage{times}
\usepackage{latexsym}

\usepackage[T1]{fontenc}

\usepackage[utf8]{inputenc}

\usepackage{microtype}

\usepackage{inconsolata}

\usepackage{graphicx}
\usepackage{multirow}

\usepackage{booktabs}
\usepackage{multirow}
\usepackage{amsmath} 
\usepackage{tabularx,array,amssymb}
\usepackage{booktabs}
\usepackage[most]{tcolorbox}
\usepackage{enumitem}
\usepackage{placeins}

\usepackage{xcolor}
\newcommand{\dgain}[1]{{\scriptsize\textcolor{teal}{$\uparrow$#1}}}
\newcommand{\ddrop}[1]{{\scriptsize\textcolor{red!70!black}{$\downarrow$#1}}}
\newcommand{\dsame}{{\scriptsize\textcolor{gray}{$\pm$0.0}}}
\newcommand{\tabscale}{0.72}
\newcommand{\ph}[1]{\textcolor{blue!70!black}{#1}}
\newcommand{\aud}[1]{\textcolor{teal!80!black}{#1}}
\newtcolorbox[auto counter]{promptbox}[2]{
  breakable, enhanced,
  colback=gray!4, colframe=gray!55, boxrule=0.5pt, arc=1pt,
  left=4pt, right=4pt, top=3pt, bottom=3pt,
  fonttitle=\bfseries\footnotesize, coltitle=black, colbacktitle=gray!18,
  title={Prompt~\thetcbcounter: #1}, label={#2},
  fontupper=\ttfamily\footnotesize,
  before upper={\frenchspacing\raggedright\setlength{\parskip}{0.6em}\setlength{\parindent}{0pt}},
}

\usepackage{xcolor}
\definecolor{findingbg}{RGB}{232,240,250}   %
\newcommand{\finding}[2]{\par\vspace{4pt}\noindent\colorbox{findingbg}{%
  \parbox{\dimexpr\linewidth-2\fboxsep\relax}{\textbf{Finding #1.} #2}}\par\vspace{4pt}}

\title{Do Audio LLMs HEAR Pain?\\Tracing Acoustic Evidence from Speech to Language Reasoning}

\author{
  Wenqi Li\thanks{\ \ Equal contribution.} \and
  Xiangjue Dong\footnotemark[1] \and
  Sulaiman Vesal \\
  Microsoft \\
  \texttt{\{rubyli, xiangjuedong, svesal\}@microsoft.com}
}

\begin{document}
\maketitle
\begin{abstract}
Large audio language models (LALMs) can process speech directly, but it remains unclear how well they use the non-lexical acoustic information that speech carries. We study this question through pain, using TAME-Pain, which contains pain-neutral Harvard sentences and spoken pain ratings recorded at different self-reported pain levels. 
We introduce HEAR, a four-stage diagnostic framework that traces pain-related information from the speech signal to the model's prediction, and apply it to seven closed-source and open-weight LALMs. On the Harvard sentences, where pain can only be inferred from the voice, all models perform at chance level, and same-speaker references or few-shot examples improve balanced accuracy by no more than six points. On the spoken ratings, the same models reach 87.5--99.7\% balanced accuracy.
Layer-wise probing of open-weight LALMs shows that pain can be decoded from early audio representations (74.1\% balanced accuracy for binary prediction and 45.0\% for three-class severity), but that this information, along with most pain-associated acoustic cues, decreases in deeper layers.
In an acoustic perception test, weaker models cannot distinguish even large changes in these cues, whereas stronger models can but still fail to infer pain. 
Finally, we recover pain-related acoustic cues and directly reinject them as structured text into the prompt, which improves pairwise pain prediction by up to 32 points. These results indicate that current LALMs encode paralinguistic information in early layers that is largely unavailable when the model produces its answer.
\end{abstract}

\begin{figure}[!ht]
  \centering
  \includegraphics[width=\columnwidth]{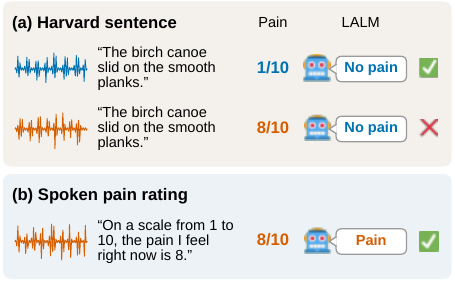}
  \caption{LALMs recognize pain when it is stated in words, but fail when it must be inferred from the voice. Examples from TAME-Pain \citep{Dao2025}, recorded during a pain-inducing cold pressor task. \textbf{(a)} Speakers read fixed, pain-neutral Harvard sentences, so pain is detectable only from the acoustic signal. \textbf{(b)} Speakers state their pain rating aloud, conveying pain both lexically and acoustically.}
  \label{fig:teaser}
  \vspace{-15pt}
\end{figure}

\section{Introduction}
Large audio language models (LALMs) are increasingly used as general speech interfaces, including in medical consultation and clinical decision support \citep{chen-etal-2026-speechmedassist, tam2025medvoicebias}. In such settings, \emph{how} something is said can matter as much as \emph{what} is said \citep{kim-etal-2026-clinicast}. However, LALMs are often evaluated on tasks where the words alone provide much of the information needed to answer correctly \citep{chen2026listen}. Strong results on these tasks therefore do not show whether a model uses the acoustic properties of speech, or mainly transcribes the input and reasons over its lexical content.
 
Pain is a clear case where this difference matters. Pain changes the voice, for example its loudness and pitch \citep{PMIDphoneicpain, gonzales2025speech}, and the voice can signal pain even when the words do not mention it. A model that relies on the words would miss exactly these cases. Prior work shows that LALMs rely mostly on lexical content for paralinguistic tasks such as emotion recognition \citep{chen2026listen,pang2026voxparadox}. It is still unclear why: is the acoustic information missing from the model, lost
inside it, or present but not used?

\begin{figure*}[!ht]
  \centering
  \includegraphics[width=\textwidth]{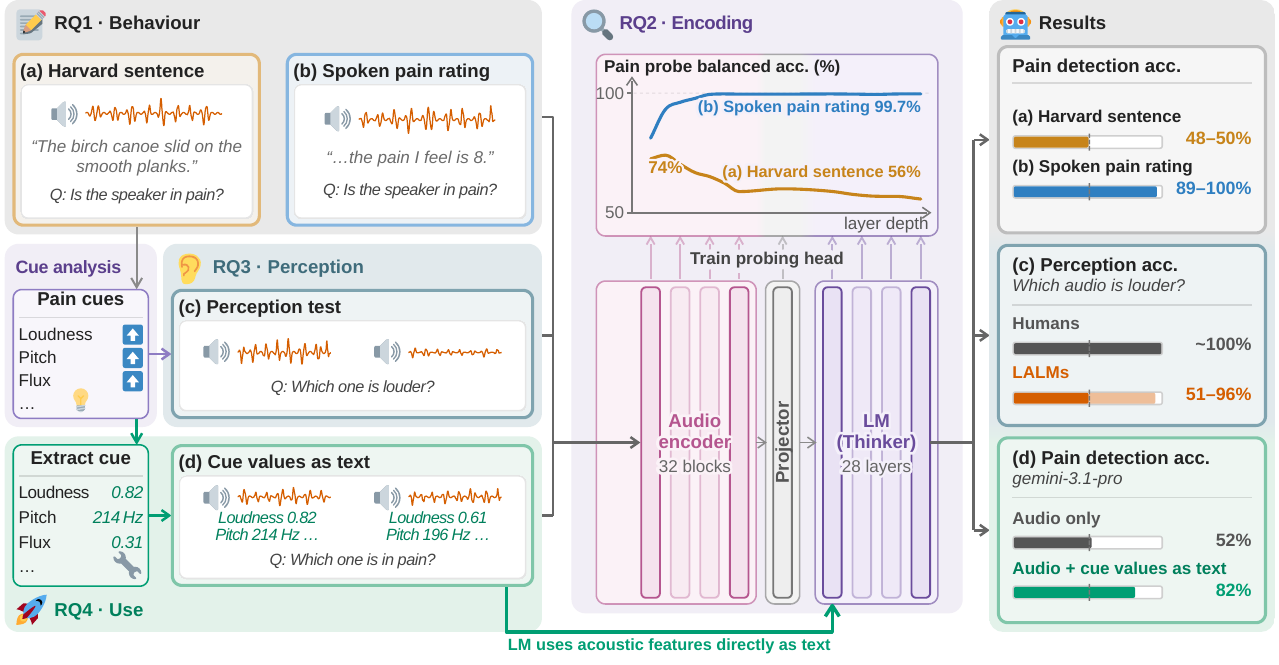}
  \caption{\textbf{Overview of our HEAR analysis framework.} \textbf{(RQ1)}: we compare seven LALMs' pain prediction under the neutral-sentence and spoken-rating settings. \textbf{(RQ2)}: we identify pain-associated acoustic cues and probe each layer of LALMs for pain representations. \textbf{(RQ3)}: we test whether LALMs perceive large, controlled contrasts in these cues. \textbf{(RQ4)}: we add acoustic cue values to the prompt as text.}
  \label{fig:probe_architecture}
  \vspace{-10pt}
\end{figure*}

We study this question with TAME-Pain \citep{Dao2025}, recorded during a pain-inducing cold pressor task (Figure~\ref{fig:teaser}).
Speakers read pain-neutral Harvard sentences \citep{harvardsent}, so pain can only be inferred from the voice. They also state their pain rating aloud, so pain is directly inferable from the words. Comparing the two conditions separates acoustic from lexical evidence without editing the speech. Building on this comparison, we propose \textbf{HEAR}, a framework that analyzes pain-related information at four stages, following it from the input signal to the model's prediction (Figure~\ref{fig:probe_architecture}):

\begin{itemize}[leftmargin=*,itemsep=0pt,topsep=0pt,parsep=0pt]
  \item \textbf{RQ1} \textbf{Behavior:} Do LALMs recognize pain when it
  is conveyed only via the acoustic signal?
  \item \textbf{RQ2} \textbf{Encoding:} Is pain-related information
  present in the audio and in the model's internal representations, and
  at which stage is it lost?
  \item \textbf{RQ3}  \textbf{Perception:} Can LALMs discriminate the
  acoustic cues associated with pain?
  \item \textbf{RQ4}  \textbf{Use:} Can LALMs reason about pain when
  these cues are provided explicitly?
\end{itemize}

Across seven closed-source and open-weight LALMs, when pain is conveyed only via the acoustic signal, models perform near chance on predicting pain. Same-speaker references and 20-shot examples provide little improvement. When the pain rating is spoken explicitly, however, the same models approach ceiling performance (RQ1).

We then examine where this failure arises (Figure~\ref{fig:probe_architecture}). Acoustic analysis shows that pain is associated with properties including loudness, pitch and spectral characteristics. In Qwen2.5-Omni-7B~\citep{xu2025qwen25omnitechnicalreport}, supervised probes recover pain information above chance from early audio representations, reaching 74.1\% balanced accuracy for binary classification and 45.0\% for three-class classification. Pain-related information becomes less decodable in deeper representations (RQ2), and the model is also near chance when directly asked to discriminate several pain-associated acoustic cues. Stronger models identify most changes but still fail to infer pain (RQ3). When the cue values are added to the prompt as text, pairwise pain prediction improves by up to 32 points (RQ4). These results suggest a gap between acoustic information represented inside the model and the information that remains accessible for downstream language-level reasoning. Our contributions are threefold:
\begin{itemize}[leftmargin=*,itemsep=0pt,topsep=0pt,parsep=0pt]
    \item We introduce HEAR, a \textbf{controlled lexical--acoustic evaluation of pain understanding in LALMs}, disentangling acoustic-lexical cues across pain-neutral utterances and explicit spoken pain reports.
    \item We show a \textbf{consistent gap between lexical and paralinguistic pain understanding in LALMs} across seven closed-source and open-weight LALMs: models perform near chance when pain must be inferred from the voice, but approach ceiling when it is explicitly stated in speech.
    \item We trace \textbf{how pain-related information and acoustic cues flow through open-weight LALMs}, showing that they are most decodable in early audio representations and become less accessible with depth. We further show that reinjecting acoustic cues as text can improve pain prediction in within-speaker comparative settings.
\end{itemize}

\section{Related Work}

\paragraph{Speech-based pain assessment.}
Prior work has explored speech as a non-invasive signal for pain, using vocal and prosodic features to predict pain from both chronic and experimentally induced conditions \citep{oshrat2016speech,ren2018pain,duey2023pain}. TAME-Pain \citep{Dao2025} extends this line of work with controlled recordings collected during a cold pressor task and paired with self-reported pain levels. Recent systems have also begun incorporating learned speech representations for pain assessment \citep{kulce2026echo}. Rather than developing another supervised pain detector, we use pain as a controlled setting to examine what general-purpose audio language models capture from the acoustic signal.

\paragraph{Paralinguistic understanding in audio language models.}
Recent work shows that strong speech recognition does not necessarily imply strong paralinguistic understanding. LISTEN \citep{chen2026listen} finds that LALMs rely heavily on lexical information for emotion recognition, while ParA-LLM \citep{anand2026parallm} documents substantial gaps across a broader set of speaker, expressive, and acoustic attributes. VoxParadox \citep{pang2026voxparadox} uses controlled lexical--acoustic conflicts and layer-wise probing to study where paralinguistic information becomes less decodable. Closely related, \citet{koduru2026heard} find that speaking-style information can be encoded yet degraded before reaching LALMs' outputs. Other work explores improving this capability through targeted training and reinforcement learning \citep{chen2026paralinguistic}. Our work studies this problem using recorded human speech under experimentally induced pain and traces both pain-related information and associated acoustic cues through open-weight LALMs.

\paragraph{Interpreting audio language models.}
Interpretability work has begun to examine the internal representations of LALMs. AR\&D \citep{chowdhury2026ard} uses sparse autoencoders to identify and steer interpretable audio concepts, while \citet{fan2026readmind} use logit-lens and causal interventions to study how audio-derived information develops across layers. Our analysis is complementary: rather than searching broadly for interpretable concepts, we trace a specific paralinguistic signal from measurable acoustic cues to pain representations and downstream model behavior.

\section{Methodology}

\begin{figure*}[!ht]
\centering
\includegraphics[width=\textwidth]{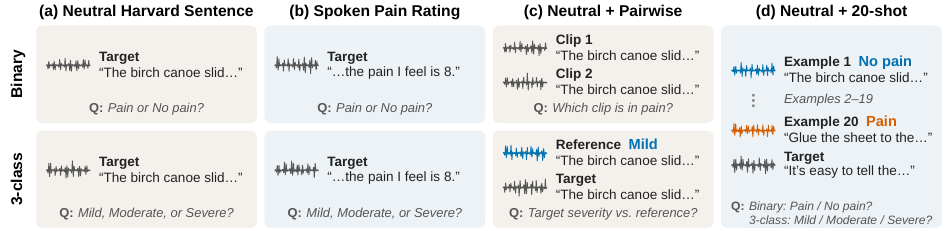}
\caption{\textbf{RQ1 pain prediction task settings}. Each recording is presented as audio, its transcript, or both. The figure is a schematic illustration and questions are simplified. The exact prompts are given in Appendix~\ref{app:prompts}.}
\label{fig:tasks}
\end{figure*}

\paragraph{Analysis framework.} Our framework HEAR (Figure~\ref{fig:probe_architecture}) uses a lexical--acoustic contrast: the same speakers, under the same pain conditions, produce speech in which pain is either absent from the words (neutral Harvard sentences) or explicitly stated (spoken pain ratings). Any difference in model behavior between the two conditions can therefore be attributed to the evidence source rather than to the speakers or the label. Building on this contrast, we trace pain-related information through four stages.
We first measure whether models recognize pain from each condition (\textbf{behavior}, \S\ref{sec:rq1}). We then test whether the signal is present in the audio and in the model's internal representations, and at which depth it is lost (\textbf{encoding}, \S\ref{sec:rq2}). Next, we test whether models can discriminate the acoustic cues associated with pain when asked directly (\textbf{perception}, \S\ref{sec:rq3}). 
Finally, we provide these cues explicitly as text to test whether models can reason over them (\textbf{use}, \S\ref{sec:rq4}).

\paragraph{Dataset and preprocessing.}
TAME-Pain \citep{Dao2025} pairs adult speech with acute pain induced by a cold pressor task (in which participants submerge a hand in 0--4\,$^\circ$C water), plus a warm-water control condition. It contains two recording types: lexically pain-neutral Harvard sentence readings (Figure~\ref{fig:teaser}a), so any pain signal in them must be acoustic, and spoken pain ratings (Figure~\ref{fig:teaser}b). Each sample carries a participant-reported, revised 1--10 pain level. Given the limited dataset size, we pool the cold and warm conditions and drop samples flagged as noisy, retaining 4,658 recordings from 51 speakers. We also use the neutral-text subset of LISTEN~\citep{chen2026listen} as an additional dataset to test whether our findings generalize.
For acoustic analysis, we extract 88 eGeMAPSv02 features \citep{7160715} with openSMILE \citep{eyben2010opensmile}.%

\paragraph{Task settings.} \label{sec:tasks}

We formulate each task as a multiple-choice question: given the audio, its transcript, or both, the model classifies the speaker's pain level (Figure~\ref{fig:tasks}) from randomly shuffled choices. Following TAME-Pain, discrete self-reported ratings are mapped to binary labels (\textit{no pain}: 1--3; \textit{pain}: 4--10) and three-class labels (\textit{mild}: 1--3; \textit{moderate}: 4--6; \textit{severe}: 7--10). For each label scheme, we construct 10 tasks spanning four settings and three input modalities (full list in Table~\ref{tab:tasks}, Appendix~\ref{app:task_desp}; prompts in Appendix~\ref{app:prompts}):
\begin{itemize}
[leftmargin=*,itemsep=1pt,topsep=2pt]
    \item \textbf{Neutral Harvard Sentence} (Figure~\ref{fig:tasks}a): predict pain from lexically neutral speech.
    \item \textbf{Spoken Pain Rating} (Figure~\ref{fig:tasks}b): predict the pain from verbalized pain rating audio.
    \item \textbf{Neutral + Pairwise} (Figure~\ref{fig:tasks}c): two clips of the same speaker reading the same sentence. Binary: pick the clip more likely in pain; three-class: rate the target's severity given a \textit{mild} reference.
    \item \textbf{Neutral + 20-shot} (Figure~\ref{fig:tasks}d): 20 labeled clips and the target, all from the same speaker. Examples are label-balanced for binary but not for three-class, due to data limits.
\end{itemize}

\paragraph{Models.} We primarily evaluate seven LALMs: four closed-source models namely gpt-realtime-mini~\citep{openai2025gptrealtimemini}, gpt-realtime~\citep{openai2025gptrealtime}, gemini-3.5-flash~\citep{gemini35flash2026}, and gemini-3.1-pro-preview~\citep{gemini31pro2026}; and three open-weight models namely Qwen2.5-Omni-7B, Qwen3-Omni-30B-A3B-Instruct~ \citep{xu2025qwen3omnitechnicalreport}, and Kimi-Audio-7B-Instruct~\citep{kimiteam2025kimiaudiotechnicalreport}. Model description and decoding settings are listed in Table~\ref{tab:models} in Appendix~\ref{app:models}.

\paragraph{Supervised probing.}
To test whether pain-related information is present in the model's representations, we train supervised probing heads for the binary and three-class tasks as defined in Section~\ref{sec:tasks}, using frozen hidden states of open-weight LALMs on the neutral Harvard sentence subset and spoken pain rating subset separately. We extract features from multiple sampled layers including audio encoder, projector and Language Model (LM) layers. We compare linear, MLP, and CRNN \citep{akir2017ConvolutionalRN} heads, and select the layer and head type for each task by speaker-grouped 5-fold cross-validation on the training speakers. Test speakers are held out from all training and model selection. For all supervised probing, we use the same speaker-independent split with 41 training and 10 held-out test speakers (Table~\ref{tab:test-speakers}) to prevent speaker-level leakage. Because prior supervised work on this dataset uses a different split and hyperparameter tuning setting, its results are not directly comparable to ours. Details are given in Appendix~\ref{app:prob_setting}.

\paragraph{Evaluation.}
We primarily report balanced accuracy because of imbalanced pain labels. The text-only conditions serve two purposes: to test whether the model can infer pain from verbal content on spoken pain rating; to test whether the model hallucinates pain on neutral Harvard sentences, which contain no lexical evidence of pain (so we report accuracy against a control label of \textit{no pain}).

\begin{table*}[!ht]
\centering
\resizebox{\textwidth}{!}{%
\begin{tabular}{@{}cl ccc ccc cc cc@{}}
\toprule
& & \multicolumn{3}{c}{Neutral Harvard Sentence}
& \multicolumn{3}{c}{Spoken Pain Rating}
& \multicolumn{2}{c}{Neutral + Pairwise}
& \multicolumn{2}{c}{Neutral + 20-shot} \\
\cmidrule(lr){3-5}\cmidrule(lr){6-8}\cmidrule(lr){9-10}\cmidrule(lr){11-12}
& Model & Text$^{*}$ & Audio & Text+Audio
      & Text & Audio & Text+Audio
      & Audio & Text+Audio
      & Audio & Text+Audio \\
\midrule
\multicolumn{12}{c}{\textbf{Pain vs.\ No Pain (Balanced Accuracy, \%)}} \\
\midrule
\multirow{4}{*}{\begin{tabular}{@{}c@{}}\textit{Closed-}\\\textit{Weight}\end{tabular}}
& gpt-realtime-mini      & 92.7 & 48.1 & 50.2 & 98.9 & 98.7 & 98.7 & 48.6 & 45.2 & \underline{52.8} & \underline{50.4} \\
& gpt-realtime           & 93.7 & 49.6 & \textbf{50.7} & 98.7 & 98.9 & 98.9 & 50.0 & 50.0 & \textbf{54.0} & 48.5 \\
& gemini-3.5-flash       & \underline{99.1} & 49.9 & 50.0 & 98.8 & 99.2 & \underline{99.2} & 40.9 & 43.3 & 47.6 & 47.2 \\
& gemini-3.1-pro-preview & \textbf{99.5} & \underline{50.0} & 50.0 & \underline{99.0} & \textbf{99.7} & \textbf{99.4} & \textbf{52.4} & 49.5 & 49.8 & \textbf{51.9} \\
\cmidrule(lr){1-12}
\multirow{3}{*}{\begin{tabular}{@{}c@{}}\textit{Open-}\\\textit{Weight}\end{tabular}}
& Qwen2.5-Omni-7B        & 96.6 & 49.8 & 50.0 & 98.8 & 98.8 & 98.6 & 50.5 & \textbf{52.9} & 47.7 & 48.9 \\
& Qwen3-Omni-30B-A3B-Instruct & 97.3 & \textbf{50.2} & 50.1 & \textbf{99.1} & \underline{99.5} & 99.1 & \underline{51.9} & \underline{51.0} & 49.1 & 47.6 \\
& Kimi-Audio-7B-Instruct & --   & \underline{50.0} & \underline{50.5} & --   & 89.1 & 77.0 & 49.0 & 46.2 & 51.3 & 48.5 \\
\midrule
\multicolumn{12}{c}{\textbf{3-Class Pain Severity (Balanced Accuracy, \%)}} \\
\midrule
\multirow{4}{*}{\begin{tabular}{@{}c@{}}\textit{Closed-}\\\textit{Weight}\end{tabular}}
& gpt-realtime-mini      & 89.3 & \textbf{33.7} & 33.5 & \underline{99.3} & 96.0 & 93.6 & 33.2 & \textbf{33.6} & \textbf{38.9} & 32.1 \\
& gpt-realtime           & 82.9 & 32.8 & \textbf{34.1} & \textbf{99.4} & 98.9 & \textbf{99.3} & \textbf{35.1} & 33.1 & 35.0 & 31.3 \\
& gemini-3.5-flash       & \textbf{99.2} & 33.3 & 33.4 & 99.2 & \textbf{99.5} & \underline{99.1} & 33.1 & 31.6 & 35.2 & 35.9 \\
& gemini-3.1-pro-preview & \underline{99.0} & 33.2 & 33.3 & 99.2 & \underline{99.2} & \textbf{99.3} & 33.1 & 31.7 & \underline{36.1} & \underline{38.6} \\
\cmidrule(lr){1-12}
\multirow{3}{*}{\begin{tabular}{@{}c@{}}\textit{Open-}\\\textit{Weight}\end{tabular}}
& Qwen2.5-Omni-7B        & 91.3 & \underline{33.4} & 33.1 & 96.4 & 98.0 & 93.3 & \underline{33.3} & \underline{33.3} & 29.4 & 32.0 \\
& Qwen3-Omni-30B-A3B-Instruct & 96.0 & \underline{33.4} & 33.3 & 99.1 & \textbf{99.5} & 97.8 & 32.5 & 33.0 & 34.6 & 35.5 \\
& Kimi-Audio-7B-Instruct & --   & \textbf{33.7} & \underline{34.0} & --   & 87.5 & 81.0 & \underline{33.3} & \underline{33.3} & 32.5 & \textbf{39.1} \\
\bottomrule
\end{tabular}%
}
\caption{\textbf{Results on pain detection across 7 LALMs and 4 settings}. Text, Audio, and Text+Audio denote the input modalities. All scores are balanced accuracy (\%), for which chance level is 50.0 for binary detection and 33.3 for 3-class severity, except the Neutral Harvard Sentence / Text column ($^{*}$), which reports accuracy. Because neutral transcripts carry no pain information, all text-only items are labeled ``no pain''. Kimi-Audio does not accept text-only input (–). Bold and underline mark the best and second-best results in each column.}
\label{tab:main}
\end{table*}

\section{RQ1: Do LALMs Recognize Pain from the Voice?}
\label{sec:rq1}
Table~\ref{tab:main} summarizes LALM performance across input modalities and settings (Figure~\ref{fig:tasks}). We highlight four key findings from the results:
\begin{itemize}[leftmargin=*,itemsep=1pt,topsep=2pt]
    \item \textbf{LALMs fail to detect pain from paralinguistic cues.} On neutral Harvard sentences, all LALMs perform at chance. Neither same-speaker references nor 20-shot examples help.
    \item \textbf{The models understand what is said, not how it is said.} In the spoken pain rating setting, where pain is verbalized, the same models are near ceiling from audio alone.
    \item \textbf{LALMs fail in different patterns.} Most models (except the gpt-realtime family) answer \textit{No pain} for almost every input on neutral Harvard audio per the confusion matrices (Figure~\ref{fig:confusion}). In the pairwise setting, most models pick Clip~2 regardless of content (gpt-realtime for all 208 queries). With 20 examples, \textit{Pain} answers increase (e.g., from under 1\% to 66\% for gemini-3.5-flash), but equally for pain and no-pain clips. These patterns reflect each model's default preference rather than access to pain signals in the audio.
    \item \textbf{Audio induces hallucinated pain.} Given only neutral Harvard transcripts, models rarely predict pain; given the audio of the same sentences, gpt-realtime answers \textit{Pain} for 85.0\% of no-pain clips (vs.\ 6.3\% from text alone).
\end{itemize}

\finding{1}{LALMs recognize pain from lexical content but not from the acoustic properties of speech, even when same-speaker comparisons or labeled examples are provided. Their errors reflect fixed response preferences rather than use of acoustic evidence.}

\section{RQ2: Is Pain-Related Information Encoded?}
\label{sec:rq2}

Results in RQ1 leave open whether the failure lies in the data or in the models. We distinguish these explanations by testing whether pain is reflected in the acoustics, and then whether and where this information is encoded in model representations.%

\begin{figure}[t]
\centering
\includegraphics[width=\columnwidth]{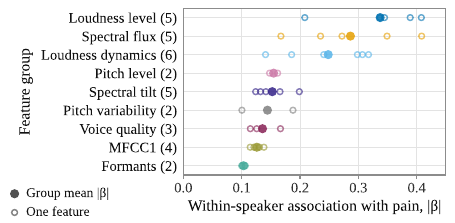}
\caption{\textbf{Acoustic feature groups associated with pain} (34 features with $q<.05$ and $|\beta|\geq.10$; groups in Table~\ref{tab:cue-groups}). $|\beta|$: within-speaker standardized association strength. Filled dots: group mean (feature number in parentheses); open circles: individual features.}
\label{fig:pain-cues}
\vspace{-2mm}
\end{figure}

\begin{figure*}[t]
  \centering
  \includegraphics[width=\textwidth]{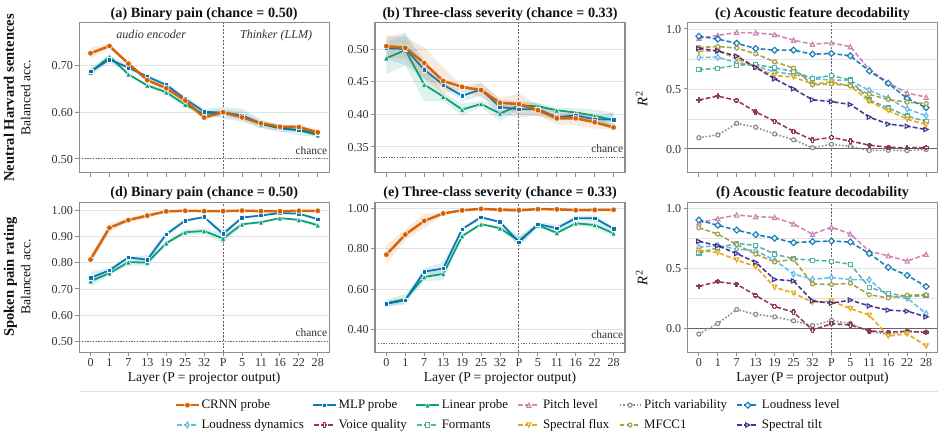}
  \caption{\textbf{Pain and acoustic-feature information across the layers of Qwen2.5-Omni-7B.} \textbf{(a, b, d, e)}~Out-of-fold balanced accuracy of probes trained on each layer's representation (speaker-grouped five-fold cross-validation; line = mean over three seeds, band = $\pm 1$~SD). Y-axis ranges differ between rows. \textbf{(c, f)}~Held-out $R^2$ (on test speakers) of ridge regression recovering the pain-associated acoustic features from the same representations, averaged within each feature group (Table~\ref{tab:cue-groups}). The projector output (P) is the input to Thinker layer~0.}
  \label{fig:layerwise}
\end{figure*}

\paragraph{Acoustic features are associated with pain.}
Using only the training speakers, we center each eGeMAPSv02 feature within speaker and fit a linear mixed-effects model regressing normalized pain ratings on each speaker-centered, standardized feature, with a random intercept per speaker. After Benjamini--Hochberg correction, 34 of the 88 features are significantly associated with pain ($q<.05$, $|\beta|\geq.10$); we group them into nine categories (Figure~\ref{fig:pain-cues}; Table~\ref{tab:cue-groups}; all coefficients in Table~\ref{tab:egemaps-association}). The strongest associations involve loudness, spectral flux, and pitch, consistent with prior evidence that pain alters vocal loudness and pitch \citep{PMIDphoneicpain, gonzales2025speech}. These features also carry usable signal: a LightGBM classifier on all 88 features, with the same split and within-speaker normalization, reaches 65.7\% balanced accuracy on binary detection and 47.1\% on three-class severity.

\paragraph{Pain information is encoded in LALM representations.}
Probes on frozen Qwen2.5-Omni-7B hidden states trained under neutral Harvard sentence setting reach 74.1\% balanced accuracy on binary detection and 45.0\% on three-class severity (Table~\ref{tab:probing}), well above the model's own near-chance answers in RQ1 (Table~\ref{tab:main}). A perturbation analysis shows that the probes rely mainly on spectral balance and pauses, and to a lesser extent on loudness and pitch (Appendix~\ref{sec:perturb_prob}). The pain signal is thus present in the model's early audio representations but does not surface in its outputs.

\begin{table}[!ht]
\centering
\small
\begin{tabular}{l cc cc}
\toprule
& \multicolumn{2}{c}{Pain vs.\ No Pain}
& \multicolumn{2}{c}{3-Class Severity} \\
\cmidrule(lr){2-3}\cmidrule(lr){4-5}
Probe & Layer & BA & Layer & BA \\
\midrule
Linear       & Block 1 & 68.8 {\scriptsize$\pm$0.8} & Block 1 & \textbf{45.0} {\scriptsize$\pm$0.9} \\
MLP          & Block 1 & 67.1 {\scriptsize$\pm$2.0} & Input   & 43.1 {\scriptsize$\pm$1.7} \\
CRNN         & Block 1 & \textbf{74.1} {\scriptsize$\pm$1.6} & Input   & 44.4 {\scriptsize$\pm$3.5} \\
\bottomrule
\end{tabular}
\caption{\textbf{Supervised probing results on Qwen2.5-Omni-7B}. Balanced accuracy (BA, \%) is reported on the 10 held-out test speakers. Input denotes the audio encoder input and Block 1 the first encoder block. Results are reported as mean $\pm$ standard deviation over three random seeds. Experiment setup is in Appendix~\ref{app:prob_setting}}
\label{tab:probing}
\end{table}

\paragraph{Acoustic pain information fades with depth.}
Figure~\ref{fig:layerwise} (a, b, d, e) shows probe accuracy across layers under two settings: neutral Harvard sentences, where pain is carried only by acoustics, and Spoken Pain Rating recordings, where the rating is stated in words (detail in Appendix~\ref{app:prob_setting}):
\begin{itemize}[leftmargin=*,itemsep=0pt,topsep=2pt]
  \item \textbf{Neutral Harvard sentences}: accuracy peaks in the first encoder blocks and declines steadily through the encoder and the LM.
  \item \textbf{Spoken ratings}: accuracy rises to near ceiling within the encoder and stays there.
\end{itemize}
Kimi-Audio shows the same pattern (Appendix~\ref{app:kimi}, Figure~\ref{fig:kimi-layerwise}a, b, d, e). These trends suggest that the audio encoder progressively transforms speech toward its semantic content: lexical pain information builds up in the spoken pain rating setting, while the acoustic pain signal decreases in the neutral Harvard sentence setting.

\paragraph{Acoustic cues weaken with depth and some are barely encoded.}

We next test how well these important pain-associated acoustic features (Figure~\ref{fig:pain-cues}) can be recovered from each layer. For each sampled layer, we pool frame-level hidden states into temporal statistics (mean, SD, and mean and SD of frame-to-frame differences) so that variability features like loudness dynamics remain recoverable. We then fit a ridge regression per feature, and report held-out $R^2$ on the 10 test speakers, averaged within each feature group (Figure~\ref{fig:layerwise}c, f; Table~\ref{tab:cue-groups}). Results are consistent across Qwen2.5 and Kimi-Audio (Figure~\ref{fig:kimi-layerwise}c, f in Appendix~\ref{app:kimi}):
\begin{itemize}[leftmargin=*,itemsep=0pt,topsep=2pt]
  \item \textbf{Pain cues fade by layer depth}: most groups are well recovered in shallow encoder layers and decline by layer depth. Even the strongest cues, loudness level and pitch level, become much less decodable inside the LM layers.
  \item \textbf{Some cues are barely encoded}: voice quality and pitch variability are among the significant pain-associated features, yet remain poorly recovered at every layer.
\end{itemize}
Both encoders are initialized from Whisper-large-v3~\citep{pmlr-whisper-radford23a, xu2025qwen25omnitechnicalreport, kimiteam2025kimiaudiotechnicalreport}, and LALM training largely focuses on transcription and content understanding. This may push models to trade acoustic cues for lexical content.

\finding{2}{Pain-related information is present in the audio and in early encoder representations. It becomes weaker toward the language model, while pain stated in words becomes stronger.}

\section{RQ3: Can LALMs Discriminate Pain-Related Cues?}
\label{sec:rq3}

RQ2 shows that acoustic pain signals are decodable from early layers by probing, but this does not mean the model can access them during inference. We therefore use a simple test to measure LALMs' acoustic perception: when two clips differ clearly in one cue, can the model tell which is higher?

\paragraph{Encoded cues are not always perceived, and perceived cues are not always used.}%

\begin{figure}[!t]
  \centering
  \includegraphics[width=\columnwidth]{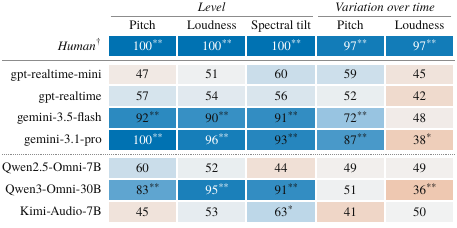}
    \caption{\textbf{LALM accuracy (\%) on the perception test.} Each item pairs two clips created from the same recording by shifting one acoustic feature in opposite directions; the model picks the clip higher on that feature (chance $=50$). $^{*}p<.05$, $^{**}p<.01$, two-sided binomial test against chance. $^{\dagger}$Human: pooled over three listeners, $n=20$ pairs per feature, Fleiss' $\kappa=0.95$.}
  \label{fig:contrast_heatmap}
  \vspace{-4mm}
\end{figure}

We construct an acoustic pain cue perception test on Harvard sentences. For five pain-associated cue groups that humans can distinguish (Figure~\ref{fig:pain-cues}), we shift the most representative features in opposite directions to create exaggerated A/B pairs (Appendix~\ref{app:contrast-dataset}; human accuracy 98.7\%). Models are asked which clip is higher on the cue, e.g., which is louder (Appendix~\ref{app:prompts-contrast}).
 Results (Figure~\ref{fig:contrast_heatmap}) show that:
\begin{itemize}[leftmargin=*,itemsep=0pt,topsep=2pt]
    \item \textbf{Frontier models perceive most cues yet do not use them:} Gemini models and Qwen3-Omni reach 72--100\% on most cues yet still fail to predict pain. Their failure on pain prediction thus suggests a deficit in reasoning over audio rather than in perception alone.
    \item \textbf{Decodable is not perceivable:} weaker models (gpt-realtime family, Qwen2.5-Omni, Kimi-Audio) stay near chance on almost every cue. Yet Qwen2.5-Omni's final LM layers still encode pitch level ($R^2\approx0.5$; Figure~\ref{fig:layerwise}c), while the model cannot tell which exaggerated clip has higher pitch. The information reaches the LM but does not surface in the output.
\end{itemize}

\finding{3}{Strong LALMs can discriminate large acoustic contrasts but still fail to use these cues for pain. Weaker models cannot discriminate them at all, even when the cues remain decodable from their final layers.}

\section{RQ4: Can LALMs Use Explicitly Provided Cues?}
\label{sec:rq4}

\begin{table*}[!ht]
\centering
\setlength{\tabcolsep}{3pt}
\scalebox{\tabscale}{%
\begin{tabular}{l ccc ccc ccc ccc}
\toprule
& \multicolumn{6}{c}{\textbf{Pain vs.\ No Pain}}
& \multicolumn{6}{c}{\textbf{3-Class Pain Severity}} \\
\cmidrule(lr){2-7}\cmidrule(lr){8-13}
& \multicolumn{3}{c}{Pairwise} & \multicolumn{3}{c}{20-shot}
& \multicolumn{3}{c}{Pairwise} & \multicolumn{3}{c}{20-shot} \\
\cmidrule(lr){2-4}\cmidrule(lr){5-7}\cmidrule(lr){8-10}\cmidrule(lr){11-13}
Model & Audio & Text+Audio & AF & Audio & Text+Audio & AF & Audio & Text+Audio & AF & Audio & Text+Audio & AF \\
\midrule
gpt-realtime-mini      & 66.3\,\dgain{17.7} & 64.9\,\dgain{19.7} & 77.4 & \textbf{63.1}\,\dgain{10.3} & 60.1\,\dgain{9.7}  & 58.8 & \textbf{40.0}\,\dgain{6.8} & \textbf{38.2}\,\dgain{4.6} & 39.1 & 36.5\,\ddrop{2.4} & 35.9\,\dgain{3.8} & 38.4 \\
gpt-realtime           & 53.8\,\dgain{3.8}  & 51.9\,\dgain{1.9}  & 67.8 & 60.4\,\dgain{6.4}  & \textbf{63.7}\,\dgain{15.2} & 60.7 & 34.6\,\ddrop{0.5} & 34.7\,\dgain{1.6} & 36.2 & 38.3\,\dgain{3.3} & 37.9\,\dgain{6.6} & \textbf{46.6} \\
gemini-3.5-flash       & 71.6\,\dgain{30.7} & 61.5\,\dgain{18.2} & 84.6 & 54.5\,\dgain{6.9}  & 53.1\,\dgain{5.9}  & 54.4 & \textbf{40.0}\,\dgain{6.9} & 36.5\,\dgain{4.9} & 38.7 & 29.4\,\ddrop{5.8} & 33.8\,\ddrop{2.1} & 37.5 \\
gemini-3.1-pro-preview & \textbf{82.2}\,\dgain{29.8} & \textbf{81.7}\,\dgain{32.2} & \textbf{89.4} & 60.7\,\dgain{10.9} & 60.6\,\dgain{8.7}  & 58.8 & 33.3\,\dgain{0.2} & 32.5\,\dgain{0.8} & \textbf{42.6} & \textbf{41.9}\,\dgain{5.8} & \textbf{42.6}\,\dgain{4.0} & 43.0 \\
\midrule
Qwen2.5-Omni-7B        & 55.8\,\dgain{5.3}  & 52.4\,\ddrop{0.5}  & 64.9 & 57.6\,\dgain{9.9}  & 58.4\,\dgain{9.5}  & \textbf{62.4} & 35.6\,\dgain{2.3} & 35.2\,\dgain{1.9} & 36.2 & 38.8\,\dgain{9.4} & 40.3\,\dgain{8.3} & 36.2 \\
Qwen3-Omni-30B         & 69.7\,\dgain{17.8} & 68.8\,\dgain{17.8} & 76.9 & 55.8\,\dgain{6.7}  & 54.2\,\dgain{6.6}  & 59.6 & 36.3\,\dgain{3.8} & 32.0\,\ddrop{1.0} & 39.5 & 37.9\,\dgain{3.3} & 39.1\,\dgain{3.6} & 40.6 \\
Kimi-Audio-7B          & 68.3\,\dgain{19.3} & 53.4\,\dgain{7.2}  & --   & 54.2\,\dgain{2.9}  & 56.4\,\dgain{7.9}  & --   & 35.5\,\dgain{2.2} & 35.2\,\dgain{1.9} & --   & 35.5\,\dgain{3.0} & 34.9\,\ddrop{4.2} & --   \\
\bottomrule
\end{tabular}%
}
\caption{\textbf{Effect of adding explicit acoustic cues to the prompt on the pairwise and 20-shot Neutral Harvard Sentence settings} (balanced accuracy, \%). Arrows give the change relative to the baseline in Table~\ref{tab:main}. AF (acoustic features only input) has no baseline. Kimi-Audio does not accept text-only input (–). Best result per column in \textbf{bold}.}
\label{tab:acoustic-cue-delta-neutral}
\end{table*}

To mitigate the encoding and perception gap discussed in Sections~\ref{sec:rq2} and~\ref{sec:rq3}, we translate acoustic features into text and include them in the prompt, so that information lost in the audio encoder can be processed directly by the LM layers.

\paragraph{Verbalized acoustic cues help.}
We attach the most correlated features from each of the 8 cue groups (formants excluded because of lower absolute coefficient) in Figure~\ref{fig:pain-cues} to the prompt, each with its raw value and an explanation, without tuning the prompt or feature engineering (Appendix~\ref{app:prompts-acoustic}). Given the low baselines in Table~\ref{tab:main}, we focus on the pairwise and 20-shot settings (Table~\ref{tab:acoustic-cue-delta-neutral}):
\begin{itemize}[leftmargin=*,itemsep=0pt,topsep=2pt]
  \item \textbf{Adding explicit cues benefits binary pairwise setting the most}: gemini-3.1-pro-preview improves by up to 32.2 points. This means when the raw cue is explicitly given in text, LALMs can reason over it and infer pain. 
  \item \textbf{Explicit pain cues do not help reasoning on audio}: performance under cues alone (AF) setting mostly outperforms cues with audio, suggesting that the benefit may come from the verbalized acoustic feature alone. And this feature does not help the model to pay attention to the raw audio. 
\end{itemize}

In contrast, under the spoken pain rating AF setting, all four evaluated models' performance drops close to the random level  (Table~\ref{tab:acoustic-cue-delta-zeroshot} in Appendix~\ref{app:cue-additional}): This again suggests that the ceiling performance with spoken pain rating is from the lexical pain information alone, not the audio, and LALMs can only infer pain when it is present verbally.

Furthermore, we observe a similar improvement trend in the pairwise setting on the LISTEN dataset~\citep{chen2026listen} (Appendix~\ref{app:cue-additional}, Table~\ref{tab:listen-pairwise}), especially for the Qwen3-Omni model.

\finding{4}{When acoustic cues are verbalized, LALMs can use them to compare pain within a speaker, but fail to do so from the audio itself. The bottleneck thus lies in a modality ability gap: models can reason over textualized cues but cannot ground pain judgments in the audio.}

\section{Discussion}

Our results point to a gap between recognizing the words in speech and using the acoustic information carried by the voice. Part of this gap arises inside the model: pain-related acoustic cues are most decodable in early audio representations and fade with depth, while lexical pain information grows stronger. Yet depth-wise loss is not the whole story. Strong models perceive most cues and some cues persist into the final layers, but neither is used to infer pain. 

Reinjecting the cues as text shows that the failure is not a lack of reasoning capacity. However, the benefit is not uniform. Gains are largest in within-speaker pairwise comparison and vanish for single recordings. Adding the audio back often lowers accuracy, suggesting that the model's own audio representation can conflict with the provided cues. Future work could test speaker-standardized rather than raw feature values, especially in zero-shot and in-context settings. Auxiliary objectives that predict acoustic features, or more diverse paralinguistic training data, could also help.

More broadly, these findings caution against treating strong speech recognition as evidence of strong speech understanding. For applications that depend on paralinguistic information, evaluation should separate lexical from acoustic evidence and examine whether task-relevant information remains accessible across the audio-language interface. Pain provides one controlled case study of this broader problem.

\section{Conclusion}
We study whether audio language models use pain-related information carried by the voice rather than by lexical content alone. Across seven LALMs, pain prediction remains near chance when lexical content is held neutral, despite strong performance when pain is explicitly verbalized. Tracing the signal through the model shows that pain-related information and several associated acoustic cues are most decodable in early audio encoder representations but become less accessible with depth. Explicitly providing recovered acoustic cues can improve prediction in same speaker comparison settings. These results indicate that current
LALMs encode paralinguistic evidence but do not make it available to
language-level reasoning, and that addressing this gap is necessary for
reliable paralinguistic understanding.

\section*{Limitations} 
Regarding dataset, TAME-Pain contains experimentally induced acute pain from 51 speakers, so the findings may not generalize to chronic pain, clinical settings, or naturally occurring speech. The dataset is relatively small, and we pool cold-pressor and warm-water recordings to increase sample size. We further complement TAME-Pain with the neutral-text subset of LISTEN \citep{chen2026listen}, where verbalized acoustic cues also improve reference-conditioned emotion recognition, which suggests that our findings extend beyond pain. Pain labels are self-reported, and acoustic correlates of pain can vary substantially across speakers, motivating our use of speaker-relative features.

Regarding experiments, although Figure~\ref{fig:contrast_heatmap} shows that gemini-3.1-pro-preview and Qwen-3-Omni-30B can differentiate between multiple audio inputs, we do not separately verify if every model can distinguish individual audio clips at the acoustic level in the multi-audio settings. Our detailed layer-wise analysis focuses primarily on Qwen2.5-Omni-7B and Kimi-Audio; although behavioral results span seven LALMs, internal representations may differ across architectures. Finally, our acoustic perturbations and textualized cue intervention simplify naturally occurring vocal changes and should be interpreted as diagnostic analyses rather than evidence that these cues alone are sufficient for real-world pain assessment.

\section*{Ethics Statement}

We used TAME-Pain under the PhysioNet Restricted Health Data License 1.5.0. All authors and data annotators who accessed the data signed the PhysioNet data use agreement. We made no attempt at re-identification and do not redistribute any audio. Proprietary models were queried via Google Cloud Vertex AI and Azure OpenAI Service, where data are not used to train or fine-tune models. 
Automated pain assessment can cause harm if it is used without clinical validation, for example if it dismisses patients whose pain is not clearly audible. Our results do not support relying on current LALMs for this purpose.

\section*{Acknowledgments}
We thank Yingcheng Cai and Yujia Lai for volunteering their time to annotate data for this study.

\bibliography{custom}

@inproceedings{eyben2010opensmile,
author = {Eyben, Florian and W{\"o}llmer, Martin and Schuller, Bj{\"o}rn},
title = {{openSMILE}: the munich versatile and fast open-source audio feature extractor},
year = {2010},
isbn = {9781605589336},
publisher = {Association for Computing Machinery},
address = {New York, NY, USA},
url = {https://doi.org/10.1145/1873951.1874246},
doi = {10.1145/1873951.1874246},
booktitle = {Proceedings of the 18th ACM International Conference on Multimedia},
pages = {1459–1462},
numpages = {4},
location = {Firenze, Italy},
series = {MM '10}
}

@ARTICLE{7160715,
  author={Eyben, Florian and Scherer, Klaus R. and Schuller, Björn W. and Sundberg, Johan and André, Elisabeth and Busso, Carlos and Devillers, Laurence Y. and Epps, Julien and Laukka, Petri and Narayanan, Shrikanth S. and Truong, Khiet P.},
  journal={IEEE Transactions on Affective Computing}, 
  title={The Geneva Minimalistic Acoustic Parameter Set ({GeMAPS}) for Voice Research and Affective Computing}, 
  year={2016},
  volume={7},
  number={2},
  pages={190-202},
  doi={10.1109/TAFFC.2015.2457417}}

@misc{kimiteam2025kimiaudiotechnicalreport,
      title={Kimi-Audio Technical Report}, 
      author={KimiTeam and Ding Ding and Zeqian Ju and Yichong Leng and Songxiang Liu and Tong Liu and Zeyu Shang and Kai Shen and Wei Song and Xu Tan and Heyi Tang and Zhengtao Wang and Chu Wei and Yifei Xin and Xinran Xu and Jianwei Yu and Yutao Zhang and Xinyu Zhou and Y. Charles and Jun Chen and Yanru Chen and Yulun Du and Weiran He and Zhenxing Hu and Guokun Lai and Qingcheng Li and Yangyang Liu and Weidong Sun and Jianzhou Wang and Yuzhi Wang and Yuefeng Wu and Yuxin Wu and Dongchao Yang and Hao Yang and Ying Yang and Zhilin Yang and Aoxiong Yin and Ruibin Yuan and Yutong Zhang and Zaida Zhou},
      year={2025},
      eprint={2504.18425},
      archivePrefix={arXiv},
      primaryClass={eess.AS},
      url={https://arxiv.org/abs/2504.18425}, 
}

@misc{xu2025qwen3omnitechnicalreport,
      title={{Qwen3-Omni} Technical Report}, 
      author={Jin Xu and Zhifang Guo and Hangrui Hu and Yunfei Chu and Xiong Wang and Jinzheng He and Yuxuan Wang and Xian Shi and Ting He and Xinfa Zhu and Yuanjun Lv and Yongqi Wang and Dake Guo and He Wang and Linhan Ma and Pei Zhang and Xinyu Zhang and Hongkun Hao and Zishan Guo and Baosong Yang and Bin Zhang and Ziyang Ma and Xipin Wei and Shuai Bai and Keqin Chen and Xuejing Liu and Peng Wang and Mingkun Yang and Dayiheng Liu and Xingzhang Ren and Bo Zheng and Rui Men and Fan Zhou and Bowen Yu and Jianxin Yang and Le Yu and Jingren Zhou and Junyang Lin},
      year={2025},
      eprint={2509.17765},
      archivePrefix={arXiv},
      primaryClass={cs.CL},
      url={https://arxiv.org/abs/2509.17765}, 
}

@article{Dao2025,
  author={Dao, Tu-Quyen and Schneiders, Eike and Williams, Jennifer and Bautista, John Robert and Seabrooke, Tina and Vigneswaran, Ganesh and Kolpekwar, Rishik and Vashistha, Ritwik and Farahi, Arya},
  title={{TAME Pain data release: Using audio signals to characterize pain}},
  journal = {Scientific Data},
  year    = {2025},
  month   = apr,
  volume  = {12},
  number  = {1},
  pages   = {595},
  doi     = {10.1038/s41597-025-04733-2},
  url     = {https://doi.org/10.1038/s41597-025-04733-2},
  issn    = {2052-4463},
    publisher={Nature Publishing Group UK, London}
}

@misc{xu2025qwen25omnitechnicalreport,
      title={{Qwen2.5-Omni} Technical Report}, 
      author={Jin Xu and Zhifang Guo and Jinzheng He and Hangrui Hu and Ting He and Shuai Bai and Keqin Chen and Jialin Wang and Yang Fan and Kai Dang and Bin Zhang and Xiong Wang and Yunfei Chu and Junyang Lin},
      year={2025},
      eprint={2503.20215},
      archivePrefix={arXiv},
      primaryClass={cs.CL},
      url={https://arxiv.org/abs/2503.20215}, 
}

@misc{openai2025gptrealtime,
  author       = {{OpenAI}},
  title = {Introducing {gpt-realtime} and {Realtime API} updates for production voice agents},
  year         = {2025},
  howpublished = {\url{https://openai.com/index/introducing-gpt-realtime/}},
  note         = {Accessed: 2026-09}
}

@misc{openai2025gptrealtimemini,
  author       = {{OpenAI}},
  title        = {gpt-realtime-mini},
  year         = {2025},
  howpublished = {\url{https://platform.openai.com/docs/models/gpt-realtime-mini}},
  note         = {Accessed: 2026-09}
}

@misc{gemini31pro2026,
  author       = {{Google DeepMind}},
  title        = {Gemini 3.1 Pro Model Card},
  year         = {2026},
  howpublished = {\url{https://deepmind.google/models/model-cards/gemini-3-1-pro/}},
  note         = {Accessed: 2026-09}
}

@misc{gemini35flash2026,
  author       = {{Google DeepMind}},
  title        = {Gemini 3.5 Flash Model Card},
  year         = {2026},
  howpublished = {\url{https://deepmind.google/models/model-cards/gemini-3-5-flash/}},
  note         = {Accessed: 2026-09}
}

@inproceedings{chen2026listen,
    title = "{{Do Audio {LLM}s Really {LISTEN}, or Just Transcribe? Measuring Lexical vs. Acoustic Emotion Cues Reliance}}",
    author = "Chen, Jingyi  and
      Guo, Zhimeng  and
      Chun, Jiyun  and
      Wang, Pichao  and
      Perrault, Andrew  and
      Elsner, Micha",
    editor = "Demberg, Vera  and
      Inui, Kentaro  and
      Marquez, Llu{\'i}s",
    booktitle = "Proceedings of the 19th Conference of the {E}uropean Chapter of the {A}ssociation for {C}omputational {L}inguistics (Volume 1: Long Papers)",
    month = mar,
    year = "2026",
    address = "Rabat, Morocco",
    publisher = "Association for Computational Linguistics",
    url = "https://aclanthology.org/2026.eacl-long.274/",
    doi = "10.18653/v1/2026.eacl-long.274",
    pages = "5848--5877",
    ISBN = "979-8-89176-380-7"
}

@article{kulce2026echo,
    title = {{ECHO}: A Locally-Deployable Agentic Health Assistant with Temporal Memory, Safety Guardrails, and Speech Assessment},
    author = {K{\"u}l{\c{c}}e, Abdulkadir and Esen, Alihan and Fikir, {\c{C}}a{\u{g}}la and Kurt, Berke and Arar, Kuzey and Ercan, G{\"o}khan and Tek, Faik Boray},
    journal = {arXiv preprint arXiv:2608.06110},
    year = {2026}
}

@inproceedings{
pang2026voxparadox,
title={Do Audio {LLM}s Listen or Read? Analyzing and Mitigating Paralinguistic Failures with VoxParadox},
author={Jiacheng Pang and Ashutosh Chaubey and Mohammad Soleymani},
booktitle={Forty-third International Conference on Machine Learning},
year={2026},
url={https://openreview.net/forum?id=v7rYbRR9Zw}
}

@inproceedings{anand2026parallm,
  title     = {{ParA-LLM: A Unified Approach to Paralinguistic and Acoustic Speech Understanding}},
  author    = {Nishit Anand and Jiaqi Su and Ke Chen and Yunyun Wang and Dinesh Manocha and Ramani Duraiswami and Rithesh Kumar and Zeyu Jin},
  year      = {2026},
  booktitle = {{Interspeech 2026}},
  pages     = {495--500},
  doi       = {10.21437/Interspeech.2026-3015},
  issn      = {2958-1796},
}

@inproceedings{
koduru2026heard,
title={Heard but Not Heeded: Paralinguistic Information Encoding and Loss in Audio-Language Models},
author={Bhuvan Chandra Koduru and Dareen Safar Alharthi and Rita Singh and Bhiksha Raj},
booktitle={Third Conference on Language Modeling},
year={2026},
url={https://openreview.net/forum?id=GKaYbVlAgJ}
}

@inproceedings{chen2026paralinguistic,
    title = "Aligning Paralinguistic Understanding and Generation in Speech {LLM}s via Multi-Task Reinforcement Learning",
    author = "Kim, Minseok  and
      Chen, Jingxiang  and
      Leem, Seong-Gyun  and
      Huang, Yin  and
      Rungta, Rashi  and
      Ouyang, Zhicheng  and
      Wu, Haibin  and
      Appini, Surya Teja  and
      Bansal, Ankur  and
      Bai, Yang  and
      Liu, Yue  and
      Metze, Florian  and
      Aly, Ahmed A  and
      Kumar, Anuj  and
      Rastrow, Ariya  and
      Lin, Zhaojiang",
    editor = {Matusevych, Yevgen  and
      Eryi{\u{g}}it, G{\"u}l{\c{s}}en  and
      Aletras, Nikolaos},
    booktitle = "Proceedings of the 19th Conference of the {E}uropean Chapter of the {A}ssociation for {C}omputational {L}inguistics (Volume 5: Industry Track)",
    month = mar,
    year = "2026",
    address = "Rabat, Morocco",
    publisher = "Association for Computational Linguistics",
    url = "https://aclanthology.org/2026.eacl-industry.49/",
    doi = "10.18653/v1/2026.eacl-industry.49",
    pages = "636--648",
    ISBN = "979-8-89176-384-5",
}

@article{fan2026readmind,
    title = {{Can We Read the Mind of an Audio LLM? A Verbalizable, Multilingual Middle-Layer Workspace}},
    author = {Fan, Jiajun and Li, Jingyuan and Shivakumar, Prashanth Gurunath and Luo, Qi and Huang, Jia-Hong and Maruf, M. and Ren, Roger and Gu, Yile and Pandey, Rahul and Liu, Ge and Bulyko, Ivan},
    journal = {arXiv preprint arXiv:2608.24958},
    year = {2026}
}

@inproceedings{oshrat2016speech,
    title = {Speech prosody as a biosignal for physical pain detection},
    author = {Oshrat, Yaniv and Bloch, Ayala and Lerner, Anat and Cohen, Azaria and Avigal, Mireille and Zeilig, Gabi},
    booktitle = {Speech Prosody 2016},
    pages = {420--424},
    year = {2016},
    doi = {10.21437/SpeechProsody.2016-86}
}

@inproceedings{ren2018pain,
    title = {Evaluation of the Pain Level from Speech: Introducing a Novel Pain Database and Benchmarks},
    author = {Ren, Zhao and Cummins, Nicholas and Han, Jing and Schnieder, Sebastian and Krajewski, Jarek and Schuller, Bj{\"o}rn W.},
    booktitle = {Speech Communication; 13th ITG-Symposium},
    pages = {1--5},
    year = {2018}
}

@inproceedings{chen-etal-2026-speechmedassist,
    title = "{S}peech{M}ed{A}ssist: Efficiently and Effectively Adapting Speech Language Models for Medical Consultation",
    author = "Chen, Sirry  and
      Wang, Jieyi  and
      Chen, Wei  and
      Wei, Zhongyu",
    editor = "Liakata, Maria  and
      Moreira, Viviane P.  and
      Zhang, Jiajun  and
      Jurgens, David",
    booktitle = "Proceedings of the 64th Annual Meeting of the {A}ssociation for {C}omputational {L}inguistics (Volume 1: Long Papers)",
    month = jul,
    year = "2026",
    address = "San Diego, California, United States",
    publisher = "Association for Computational Linguistics",
    url = "https://aclanthology.org/2026.acl-long.1428/",
    doi = "10.18653/v1/2026.acl-long.1428",
    pages = "30914--30935",
    ISBN = "979-8-89176-390-6"
}

@article{tam2025medvoicebias,
    title = {{MedVoiceBias}: A Controlled Study of Audio {LLM} Behavior in Clinical Decision-Making},
    author = {Tam, Zhi Rui and Chen, Yun-Nung},
    journal = {arXiv preprint arXiv:2511.06592},
    year = {2025}
}

@inproceedings{kim-etal-2026-clinicast,
    title = "{C}lini{CAST}: Benchmarking Acoustic Grounding and Text Dominance in Medical Triage",
    author = "Kim, Kyusik  and
      Yoo, Hyunwoo  and
      Choi, Jaehoon  and
      Kim, Kitae  and
      Rosen, Gail  and
      Suh, Bongwon",
    editor = "Liakata, Maria  and
      Moreira, Viviane P.  and
      Zhang, Jiajun  and
      Jurgens, David",
    booktitle = "Findings of the {A}ssociation for {C}omputational {L}inguistics: {ACL} 2026",
    month = jul,
    year = "2026",
    address = "San Diego, California, United States",
    publisher = "Association for Computational Linguistics",
    url = "https://aclanthology.org/2026.findings-acl.2056/",
    doi = "10.18653/v1/2026.findings-acl.2056",
    pages = "41321--41343",
    ISBN = "979-8-89176-395-1"
}

@article{duey2023pain,
    title = {Daily Pain Prediction Using Smartphone Speech Recordings of Patients With Spine Disease},
    author = {Duey, Akiro H. and Rana, Aakanksha and Siddi, Francesca and Hussein, Helweh and Onnela, Jukka-Pekka and Smith, Timothy R.},
    journal = {Neurosurgery},
    volume = {93},
    number = {3},
    pages = {670--677},
    year = {2023},
    doi = {10.1227/neu.0000000000002474}
}

@INPROCEEDINGS{chowdhury2026ard,
  author={Chowdhury, Townim Faisal and Huy, Ta Duc and Pan, Siqi and Stoddard, Jeremy and Liao, Zhibin},
  booktitle={ICASSP 2026 - 2026 IEEE International Conference on Acoustics, Speech and Signal Processing (ICASSP)}, 
  title={{AR\&D: A Framework for Retrieving and Describing Concepts for Interpreting AudioLLMs}}, 
  year={2026},
  volume={},
  number={},
  pages={1791-1795},
  doi={10.1109/ICASSP55912.2026.11460417}}

@ARTICLE{harvardsent,
author={{IEEE Subcommittee on Subjective Measurements}},
  journal={IEEE Transactions on Audio and Electroacoustics}, 
  title={{IEEE} Recommended Practice for Speech Quality Measurements}, 
  year={1969},
  volume={17},
  number={3},
  pages={225--246},
  doi={10.1109/TAU.1969.1162058}
  }

@article{akir2017ConvolutionalRN,
  title={Convolutional Recurrent Neural Networks for Polyphonic Sound Event Detection},
 author={{\c{C}}ak{\i}r, Emre and Parascandolo, Giambattista and Heittola, Toni and Huttunen, Heikki and Virtanen, Tuomas},
  journal={IEEE/ACM Transactions on Audio, Speech, and Language Processing},
  year={2017},
  volume={25},
  pages={1291-1303},
  number = {6},
  doi = {10.1109/TASLP.2017.2690575},
  url={https://api.semanticscholar.org/CorpusID:8708461}
}

@article {PMIDphoneicpain,
	Title = {Phonetic characteristics of vocalizations during pain},
	Author = {Lautenbacher, Stefan and Salinas-Ranneberg, Melissa and Niebuhr, Oliver and Kunz, Miriam},
	DOI = {10.1097/pr9.0000000000000597},
	Number = {3},
	Volume = {2},
	Month = {May},
	Year = {2017},
	Journal = {Pain reports},
	ISSN = {2471-2531},
	Pages = {e597},
	URL = {https://europepmc.org/articles/PMC5646576},
}

@article{gonzales2025speech,
  title   = {What can speech tell us about pain?},
  author  = {Gonzales, Adrian and Yao, Kylie and Vogel, Adam and Egorova-Brumley, Natalia},
  journal = {Pain Reports},
  volume  = {10},
  number  = {4},
  pages   = {e1293},
  year    = {2025},
  doi     = {10.1097/PR9.0000000000001293}
}

@article{moulines1990psola,
  author    = {Moulines, Eric and Charpentier, Francis},
  title     = {Pitch-synchronous waveform processing techniques for text-to-speech synthesis using diphones},
  journal   = {Speech Communication},
  volume    = {9},
  number    = {5--6},
  pages     = {453--467},
  year      = {1990},
  publisher = {Elsevier},
  doi       = {10.1016/0167-6393(90)90021-Z}
}

@article{Parselmouth,
  author  = {Jadoul, Yannick and Thompson, Bill and {de Boer}, Bart},
  title   = {Introducing {P}arselmouth: A {P}ython interface to {P}raat},
  journal = {Journal of Phonetics},
  volume  = {71},
  pages   = {1--15},
  year    = {2018},
  doi     = {10.1016/j.wocn.2018.07.001},
 publisher={Elsevier}
}

@misc{boersma2024praat,
  author = {Boersma, Paul and Weenink, David},
  title  = {{Praat: doing phonetics by computer}},
  note   = {Computer program},
  url    = {https://www.praat.org},
  year   = {2024}
}

@InProceedings{pmlr-whisper-radford23a,
  title = 	 {Robust Speech Recognition via Large-Scale Weak Supervision},
  author =       {Radford, Alec and Kim, Jong Wook and Xu, Tao and Brockman, Greg and Mcleavey, Christine and Sutskever, Ilya},
  booktitle = 	 {Proceedings of the 40th International Conference on Machine Learning},
  pages = 	 {28492--28518},
  year = 	 {2023},
  editor = 	 {Krause, Andreas and Brunskill, Emma and Cho, Kyunghyun and Engelhardt, Barbara and Sabato, Sivan and Scarlett, Jonathan},
  volume = 	 {202},
  series = 	 {Proceedings of Machine Learning Research},
  month = 	 {23--29 Jul},
  publisher =    {PMLR},
  url = 	 {https://proceedings.mlr.press/v202/radford23a.html}
}

\clearpage
\appendix
\section{Experiment Setting}

\subsection{Probing Experiment Setting}\label{app:prob_setting}

\paragraph{Data split.} To prevent speaker-level leakage, we hold out 10 fixed test speakers (Table~\ref{tab:test-speakers}) and train on the remaining 41 speakers. Test speakers are excluded from training, normalization, and model selection.

\begin{table}[!ht]
\centering
\small
\begin{tabular}{ccccc}
\toprule
\multicolumn{5}{c}{\textbf{Test Speaker IDs}} \\
\midrule
\texttt{p10555} & \texttt{p12460} & \texttt{p18355} & \texttt{p40505} & \texttt{p50580} \\
\texttt{p52610} & \texttt{p59520} & \texttt{p77220} & \texttt{p79550} & \texttt{p91315} \\
\bottomrule
\end{tabular}
\caption{Held-out test speakers for all supervised probing experiments on TAME-Pain dataset.}
\label{tab:test-speakers}
\end{table}

\paragraph{Features.} Hidden states are extracted from the frozen audio encoder, audio projector, and LM layers. For LM-layer features, each recording is paired with the prompt ``Listen carefully to this recording.'' to keep the hidden states task-neutral, and only audio-token positions are kept.

\paragraph{Heads.} The linear and MLP (256 hidden-dimensions) heads operate on mean-pooled features. The CRNN applies two 1D convolutions and a 2-layer bidirectional GRU over the frame-level sequence, followed by concatenated attention and mean pooling.

\paragraph{Hyperparameter Tuning.} Table~\ref{tab:probe-hparams} shows our search space during hyperparameter tuning for Qwen2.5-Omni-7B. Each fold trains for up to 30 epochs with early stopping (patience 7). The selected configuration is retrained on all training speakers for the median fold-best epoch count and evaluated once on the test speakers. Other hyperparameters are fixed: AdamW (learning rate $3\times10^{-4}$, weight decay $10^{-2}$), batch size 8, gradient clipping at 5.0, cosine schedule with 5\% warmup, and class-weighted cross-entropy. Features are standardized using statistics from the training data only.

\begin{table}[!ht]
\centering
\small
\begin{tabular}{@{}l p{5.2cm}@{}}
\toprule
Hyperparameter & Search space \\
\midrule
Layer & Encoder input; encoder blocks 1, 7, 13, 19, 25, 32; projector; LM layers 5, 11, 16, 22, 28 \\
Head & Linear, MLP, CRNN \\
\bottomrule
\end{tabular}
\caption{Pain Classification Probing hyperparameter search space for Qwen2.5-Omni-7B.}
\label{tab:probe-hparams}
\end{table}

\paragraph{Comparison with prior supervised methods.} We do not compare directly with existing supervised baselines because the settings differ: we use a speaker-independent split and we tune hyperparameters by speaker-grouped cross-validation.

\subsection{Task Description}\label{app:task_desp}
Table~\ref{tab:tasks} lists all evaluation tasks. Each task is defined by its label scheme, setting, and input modality. The corresponding prompt for each task is listed in Appendix~\ref{app:prompts}.
\begin{table*}[!ht]
\centering
\small
\begin{tabular}{l l l p{7.2cm} r}
\toprule
Prompt & Setting & Input & Description & $N$ \\
\midrule
\multicolumn{5}{l}{\textit{Pain vs.\ No Pain}} \\
P\ref{prompt:bin-zs-audio}   & Spoken Pain Rating & Audio      & Classify pain from a single recording. & 856 \\
P\ref{prompt:bin-zs-audio}   & Neutral Harvard Sentence       & Audio      & Classify pain from a single recording. & 3,802 \\
P\ref{prompt:bin-zs-text}    & Spoken Pain Rating & Text       & Classify pain from the transcript only. & 856 \\
P\ref{prompt:bin-zs-text}    & Neutral Harvard Sentence       & Text       & Classify pain from the transcript only. & 3,802 \\
P\ref{prompt:bin-zs-ta}      & Spoken Pain Rating & Text+Audio & Classify pain from a recording and its transcript. & 856 \\
P\ref{prompt:bin-zs-ta}      & Neutral Harvard Sentence       & Text+Audio & Classify pain from a recording and its transcript. & 3,802 \\
P\ref{prompt:bin-pair-audio} & Neutral + Pairwise & Audio      & Given two recordings of the same speaker reading the same sentence, choose the one recorded under pain. Each pair is shown in both orders under binary classification setting. Pairs are randomly shuffled with fixed seed under 3-class classification setting. & 208 \\
P\ref{prompt:bin-pair-ta}    & Neutral + Pairwise & Text+Audio & Same as above, with the sentence transcript provided. & 208 \\
P\ref{prompt:bin-20-audio}   & Neutral + 20-shot  & Audio      & Classify a target recording given 20 labeled same-speaker examples (10 Pain, 10 No pain). & 300 \\
P\ref{prompt:bin-20-ta}      & Neutral + 20-shot  & Text+Audio & Same as above, with transcripts for all examples and the target. & 300 \\
\midrule
\multicolumn{5}{l}{\textit{3-Class Pain Severity}} \\
P\ref{prompt:tri-zs-audio}   & Spoken Pain Rating & Audio      & Classify pain severity from a single recording. & 856 \\
P\ref{prompt:tri-zs-audio}   & Neutral Harvard Sentence       & Audio      & Classify pain severity from a single recording. & 3,802 \\
P\ref{prompt:tri-zs-text}    & Spoken Pain Rating & Text       & Classify pain severity from the transcript only. & 856 \\
P\ref{prompt:tri-zs-text}    & Neutral Harvard Sentence       & Text       & Classify pain severity from the transcript only. & 3,802 \\
P\ref{prompt:tri-zs-ta}      & Spoken Pain Rating & Text+Audio & Classify pain severity from a recording and its transcript. & 856 \\
P\ref{prompt:tri-zs-ta}      & Neutral Harvard Sentence       & Text+Audio & Classify pain severity from a recording and its transcript. & 3,802 \\
P\ref{prompt:tri-ref-audio}  & Neutral + Pairwise & Audio      & Classify the severity of a target recording given a same-speaker Mild reference recording. & 262 \\
P\ref{prompt:tri-ref-ta}     & Neutral + Pairwise & Text+Audio & Same as above, with sentence transcripts provided. & 262 \\
P\ref{prompt:tri-20-audio}   & Neutral + 20-shot  & Audio      & Classify the severity of a target recording given 20 labeled same-speaker examples. & 200 \\
P\ref{prompt:tri-20-ta}      & Neutral + 20-shot  & Text+Audio & Same as above, with transcripts for all examples and the target. & 200 \\
\midrule
\multicolumn{4}{l}{Total} & 29,888 \\
\bottomrule
\end{tabular}
\caption{Task descriptions. P$n$ refers to Prompt~$n$ in Appendix~\ref{app:prompts}. $N$ denotes the number of model queries per task}
\label{tab:tasks}
\end{table*}

\begin{table*}[!ht]
\centering
\small
\begin{tabular*}{\textwidth}{@{\extracolsep{\fill}} l l l r}
\toprule
Model & Version / Checkpoint & Decoding & Max Tokens \\
\midrule
\multicolumn{4}{l}{\textit{Closed-Source Models}} \\
gpt-realtime-mini \citep{openai2025gptrealtimemini} & API (Sep.\ 2026) & default & 1024 \\
gpt-realtime \citep{openai2025gptrealtime}          & API (Sep.\ 2026) & default & 1024 \\
gemini-3.5-flash \citep{gemini35flash2026}          & API (Sep.\ 2026) & $T=0$   & 4096 \\
gemini-3.1-pro-preview \citep{gemini31pro2026}      & API (Sep.\ 2026) & $T=0$   & 1024 \\
\midrule
\multicolumn{4}{l}{\textit{Open-Weight Models}} \\
Qwen2.5-Omni-7B \citep{xu2025qwen25omnitechnicalreport}    & \texttt{Qwen/Qwen2.5-Omni-7B} & $T=0$ & 1024 \\
Qwen3-Omni-30B-A3B \citep{xu2025qwen3omnitechnicalreport}  & \texttt{Qwen/Qwen3-Omni-30B-A3B-Instruct} & $T=0$ & 1024 \\
Kimi-Audio \citep{kimiteam2025kimiaudiotechnicalreport}    & \texttt{moonshotai/Kimi-Audio-7B-Instruct} & $T=0$ & 1024 \\
\bottomrule
\end{tabular*}
\caption{Evaluated LALMs. The Azure gpt-realtime API does not expose decoding hyperparameters or random seeds, so we use its default settings. All other models use temperature $T=0$. gemini-3.5-flash uses a larger output limit because shorter limits frequently produced truncated responses.}
\label{tab:models}
\end{table*}

\subsection{Models and Data Processing}\label{app:models}

Table~\ref{tab:models} lists the evaluated models, their versions, and decoding settings. All Gemini models were accessed through Google Cloud Vertex AI, and all GPT models through Azure OpenAI Service. For GPT models, audio is resampled to 24~kHz 16-bit PCM as the API requires, with no further normalization.

We parse each model response by taking its first non-whitespace character as the answer option. Unparseable responses are counted as incorrect. Parsing success rates are reported in Table~\ref{tab:parsing} (main results) and Table~\ref{tab:parsing-acoustic} (cue-augmented prompts).

\section{Probing on Kimi-Audio}
\label{app:kimi}

\begin{figure*}[t]
  \centering
  \includegraphics[width=\textwidth]{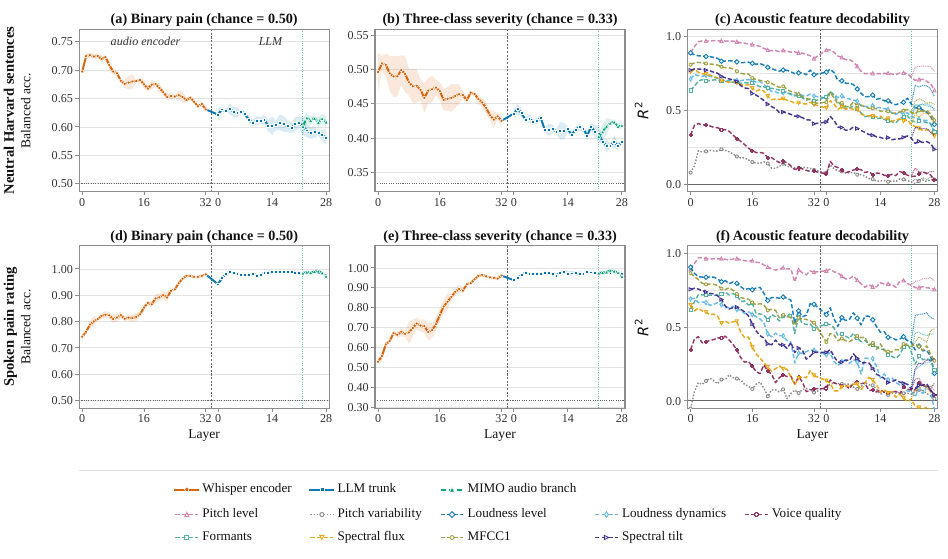}
  \caption{Layer-wise pain decoding with a linear head in Kimi-Audio. Balanced accuracy of the probe at each layer of the Whisper encoder, the LM trunk (text path), and the MIMO audio branch, which is drawn from the trunk layer it branches off (dotted green line). Lines show the mean over 3 seeds (each averaged over 5 folds); shaded bands show $\pm 1$ SD across seeds.
  (a) Binary pain detection. (b) Three-class severity.}
  \label{fig:kimi-layerwise}
\end{figure*}

To test whether the layer-wise trend generalizes beyond Qwen2.5-Omni-7B, we repeat the probing analysis with a single linear probing head on Kimi-Audio, whose LM branches into a text path and a parallel audio-token path (MIMO layers) partway up the network (Figure~\ref{fig:kimi-layerwise}).

Because all three probe heads showed the same layer-wise trend on Qwen2.5-Omni-7B (Figure~\ref{fig:layerwise}), and the CRNN is substantially more expensive to train, we use only the linear head for Kimi-Audio.

\begin{table*}[!ht]
\centering
\resizebox{\textwidth}{!}{%
\begin{tabular}{l cc cc cc cc cc cc cc}
\toprule
& \multicolumn{2}{c}{gpt-realtime-mini}
& \multicolumn{2}{c}{gpt-realtime}
& \multicolumn{2}{c}{gemini-3.5-flash}
& \multicolumn{2}{c}{gemini-3.1-pro-preview}
& \multicolumn{2}{c}{Qwen2.5-Omni-7B}
& \multicolumn{2}{c}{Qwen3-Omni-30B-A3B}
& \multicolumn{2}{c}{Kimi-Audio} \\
\cmidrule(lr){2-3}\cmidrule(lr){4-5}\cmidrule(lr){6-7}\cmidrule(lr){8-9}\cmidrule(lr){10-11}\cmidrule(lr){12-13}\cmidrule(lr){14-15}
Modality & Binary & 3-class & Binary & 3-class & Binary & 3-class & Binary & 3-class
         & Binary & 3-class & Binary & 3-class & Binary & 3-class \\
\midrule
\multicolumn{15}{l}{\textit{Neutral Harvard Sentence}} \\
Text       & 100.0 & 99.8  & 100.0 & 100.0 & 100.0 & 100.0 & 100.0 & 99.9  & 100.0 & 100.0 & 100.0 & 100.0 & --    & --    \\
Audio      & 97.3  & 98.7  & 100.0 & 100.0 & 99.4  & 99.8  & 100.0 & 99.8  & 100.0 & 100.0 & 100.0 & 100.0 & 100.0 & 100.0 \\
Text+Audio & 100.0 & 100.0 & 100.0 & 100.0 & 99.8  & 100.0 & 100.0 & 100.0 & 100.0 & 100.0 & 100.0 & 100.0 & 100.0 & 100.0 \\
\midrule
\multicolumn{15}{l}{\textit{Spoken Pain Rating}} \\
Text       & 99.8  & 100.0 & 99.9  & 100.0 & 100.0 & 100.0 & 100.0 & 100.0 & 100.0 & 100.0 & 100.0 & 100.0 & --    & --    \\
Audio      & 100.0 & 100.0 & 100.0 & 100.0 & 99.8  & 100.0 & 100.0 & 100.0 & 100.0 & 100.0 & 100.0 & 100.0 & 100.0 & 100.0 \\
Text+Audio & 100.0 & 100.0 & 100.0 & 100.0 & 100.0 & 99.9  & 100.0 & 100.0 & 100.0 & 100.0 & 100.0 & 100.0 & 100.0 & 100.0 \\
\midrule
\multicolumn{15}{l}{\textit{Neutral + Pairwise}} \\
Audio      & 100.0 & 99.6  & 100.0 & 100.0 & 100.0 & 98.9  & 100.0 & 100.0 & 100.0 & 100.0 & 100.0 & 100.0 & 100.0 & 100.0 \\
Text+Audio & 100.0 & 100.0 & 100.0 & 100.0 & 98.6  & 99.2  & 100.0 & 100.0 & 100.0 & 100.0 & 100.0 & 100.0 & 100.0 & 100.0 \\
\midrule
\multicolumn{15}{l}{\textit{Neutral + 20-shot}} \\
Audio      & 99.3  & 100.0 & 100.0 & 100.0 & 95.7  & 98.5  & 99.7  & 100.0 & 100.0 & 100.0 & 100.0 & 100.0 & 100.0 & 100.0 \\
Text+Audio & 100.0 & 100.0 & 100.0 & 100.0 & 96.0  & 96.0  & 100.0 & 100.0 & 100.0 & 100.0 & 100.0 & 100.0 & 100.0 & 100.0 \\
\midrule
Overall    & 99.6  & 99.8  & 100.0 & 100.0 & 99.0  & 99.2  & 100.0 & 100.0 & 100.0 & 100.0 & 100.0 & 100.0 & 100.0 & 100.0 \\
\bottomrule
\end{tabular}%
}
\caption{Parsing success rate (\%) of model outputs. A response counts as successfully parsed if it contains exactly one valid label. Unparsed responses are counted as incorrect in Table~\ref{tab:main}.}
\label{tab:parsing}
\end{table*}

\begin{figure*}[!ht]
  \centering
  \includegraphics[width=\textwidth]{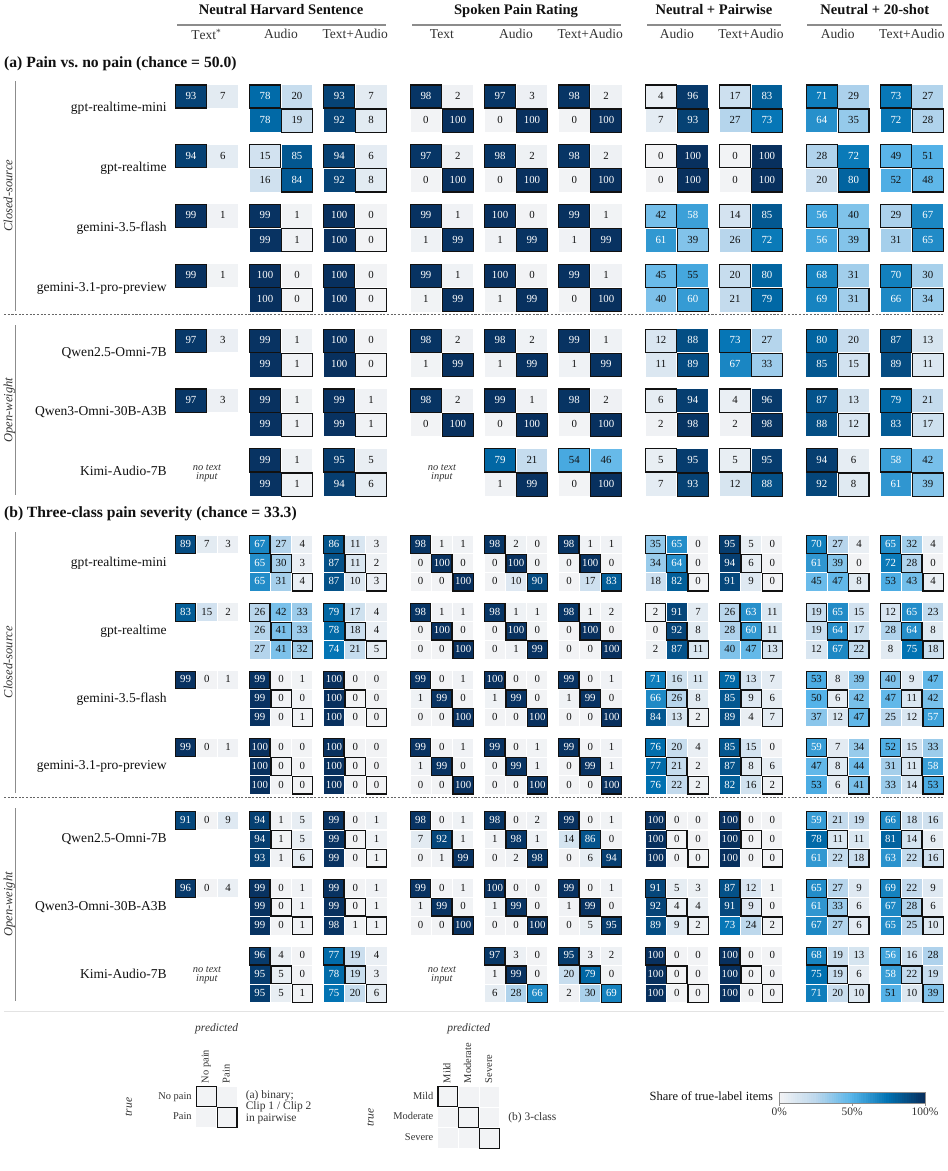}
  \caption{Confusion matrices for every model and condition in Table~\ref{tab:main}.
  Rows are true labels and columns are predicted labels; each cell gives the percentage of
  items with that true label, and outlined cells are correct predictions. Unparsed outputs
  (Table~\ref{tab:parsing}) are not shown but remain in the denominator, so a row may sum
  to slightly less than 100. In the pairwise condition, the binary labels are Clip~1 and
  Clip~2. $^{*}$All Neutral Harvard Sentence text-only items are labeled No pain (binary)
  or Mild (three-class), since neutral transcripts carry no pain information.}
  \label{fig:confusion}
\end{figure*}

\section{Additional Main Results}
\label{app:main-additional}

Table~\ref{tab:parsing} reports the parsing success rate for every condition in Table~\ref{tab:main}. All models parse at least 95.7\% of outputs in every condition and 99.0\% overall, so the low scores in the neutral-sentence settings are not explained by formatting failures.

\begin{table*}[!ht]
\centering
\scriptsize
\setlength{\tabcolsep}{3pt}
\resizebox{\textwidth}{!}{%
\begin{tabular}[t]{l r c r}
\toprule
Feature & $\beta$ & 95\% CI & $q$ \\
\midrule
\multicolumn{4}{l}{\textit{Frequency}} \\
\texttt{F0semitoneFrom27.5Hz\_sma3nz\_amean}$^{1}$ & 0.05 & [0.01, 0.08] & .011 \\
\texttt{F0semitoneFrom27.5Hz\_sma3nz\_stddevNorm} & \textbf{0.19} & [0.15, 0.22] & $<$.001 \\
\texttt{F0semitoneFrom27.5Hz\_sma3nz\_percentile20.0} & 0.00 & [$-$0.04, 0.03] & .846 \\
\texttt{F0semitoneFrom27.5Hz\_sma3nz\_percentile50.0} & \textbf{0.16} & [0.13, 0.20] & $<$.001 \\
\texttt{F0semitoneFrom27.5Hz\_sma3nz\_percentile80.0} & \textbf{0.15} & [0.11, 0.18] & $<$.001 \\
\texttt{F0semitoneFrom27.5Hz\_sma3nz\_pctlrange0-2} & \textbf{0.10} & [0.07, 0.13] & $<$.001 \\
\texttt{F0semitoneFrom27.5Hz\_sma3nz\_meanRisingSlope} & 0.05 & [0.02, 0.08] & .005 \\
\texttt{F0semitoneFrom27.5Hz\_sma3nz\_stddevRisingSlope} & 0.05 & [0.02, 0.09] & .004 \\
\texttt{F0semitoneFrom27.5Hz\_sma3nz\_meanFallingSlope} & 0.01 & [$-$0.03, 0.04] & .779 \\
\texttt{F0semitoneFrom27.5Hz\_sma3nz\_stddevFallingSlope} & 0.00 & [$-$0.04, 0.03] & .959 \\
\texttt{jitterLocal\_sma3nz\_amean}$^{2}$ & 0.01 & [$-$0.02, 0.04] & .659 \\
\texttt{jitterLocal\_sma3nz\_stddevNorm} & 0.04 & [0.01, 0.07] & .027 \\
\texttt{F1frequency\_sma3nz\_amean}$^{3}$ & 0.09 & [0.06, 0.12] & $<$.001 \\
\texttt{F1frequency\_sma3nz\_stddevNorm} & \textbf{0.11} & [0.07, 0.14] & $<$.001 \\
\texttt{F2frequency\_sma3nz\_amean} & 0.05 & [0.02, 0.08] & .006 \\
\texttt{F2frequency\_sma3nz\_stddevNorm} & 0.03 & [$-$0.01, 0.06] & .163 \\
\texttt{F3frequency\_sma3nz\_amean} & 0.04 & [0.01, 0.08] & .020 \\
\texttt{F3frequency\_sma3nz\_stddevNorm} & 0.02 & [$-$0.01, 0.05] & .331 \\
\texttt{F1bandwidth\_sma3nz\_amean}$^{4}$ & $-$0.01 & [$-$0.04, 0.02] & .635 \\
\texttt{F1bandwidth\_sma3nz\_stddevNorm} & 0.00 & [$-$0.03, 0.03] & .974 \\
\texttt{F2bandwidth\_sma3nz\_amean} & 0.00 & [$-$0.04, 0.03] & .871 \\
\texttt{F2bandwidth\_sma3nz\_stddevNorm} & 0.01 & [$-$0.02, 0.04] & .641 \\
\texttt{F3bandwidth\_sma3nz\_amean} & $-$0.04 & [$-$0.07, $-$0.01] & .035 \\
\texttt{F3bandwidth\_sma3nz\_stddevNorm} & 0.01 & [$-$0.02, 0.05] & .521 \\
\midrule
\multicolumn{4}{l}{\textit{Energy / Amplitude}} \\
\texttt{loudness\_sma3\_amean}$^{5}$ & \textbf{0.41} & [0.38, 0.44] & $<$.001 \\
\texttt{loudness\_sma3\_stddevNorm} & \textbf{0.14} & [0.11, 0.17] & $<$.001 \\
\texttt{loudness\_sma3\_percentile20.0} & \textbf{0.21} & [0.18, 0.24] & $<$.001 \\
\texttt{loudness\_sma3\_percentile50.0} & \textbf{0.34} & [0.31, 0.37] & $<$.001 \\
\texttt{loudness\_sma3\_percentile80.0} & \textbf{0.34} & [0.31, 0.38] & $<$.001 \\
\texttt{loudness\_sma3\_pctlrange0-2} & \textbf{0.31} & [0.28, 0.34] & $<$.001 \\
\texttt{loudness\_sma3\_meanRisingSlope} & \textbf{0.30} & [0.27, 0.33] & $<$.001 \\
\texttt{loudness\_sma3\_stddevRisingSlope} & \textbf{0.19} & [0.15, 0.22] & $<$.001 \\
\texttt{loudness\_sma3\_meanFallingSlope} & \textbf{0.32} & [0.29, 0.35] & $<$.001 \\
\texttt{loudness\_sma3\_stddevFallingSlope} & \textbf{0.24} & [0.21, 0.27] & $<$.001 \\
\texttt{shimmerLocaldB\_sma3nz\_amean}$^{6}$ & 0.04 & [0.01, 0.07] & .034 \\
\texttt{shimmerLocaldB\_sma3nz\_stddevNorm} & 0.03 & [0.00, 0.07] & .066 \\
\texttt{HNRdBACF\_sma3nz\_amean}$^{7}$ & 0.00 & [$-$0.03, 0.04] & .919 \\
\texttt{HNRdBACF\_sma3nz\_stddevNorm} & \textbf{0.11} & [0.08, 0.15] & $<$.001 \\
\texttt{equivalentSoundLevel\_dBp}$^{8}$ & \textbf{0.39} & [0.36, 0.42] & $<$.001 \\
\midrule
\multicolumn{4}{l}{\textit{Temporal}} \\
\texttt{loudnessPeaksPerSec}$^{9}$ & $-$0.03 & [$-$0.06, 0.01] & .149 \\
\texttt{VoicedSegmentsPerSec}$^{10}$ & 0.09 & [0.06, 0.13] & $<$.001 \\
\texttt{MeanVoicedSegmentLengthSec}$^{11}$ & 0.00 & [$-$0.03, 0.03] & .974 \\
\texttt{StddevVoicedSegmentLengthSec} & 0.01 & [$-$0.02, 0.04] & .635 \\
\texttt{MeanUnvoicedSegmentLength} & $-$0.10 & [$-$0.13, $-$0.06] & $<$.001 \\
\texttt{StddevUnvoicedSegmentLength} & $-$0.04 & [$-$0.08, $-$0.01] & .022 \\
\bottomrule
\end{tabular}%
\hspace{1.5em}%
\begin{tabular}[t]{l r c r}
\toprule
Feature & $\beta$ & 95\% CI & $q$ \\
\midrule
\multicolumn{4}{l}{\textit{Spectral balance}} \\
\texttt{spectralFlux\_sma3\_amean}$^{12}$ & \textbf{0.41} & [0.38, 0.44] & $<$.001 \\
\texttt{spectralFlux\_sma3\_stddevNorm} & \textbf{0.24} & [0.20, 0.27] & $<$.001 \\
\texttt{mfcc1\_sma3\_amean}$^{13}$ & \textbf{$-$0.14} & [$-$0.17, $-$0.10] & $<$.001 \\
\texttt{mfcc1\_sma3\_stddevNorm} & \textbf{0.13} & [0.10, 0.16] & $<$.001 \\
\texttt{mfcc2\_sma3\_amean} & $-$0.02 & [$-$0.06, 0.01] & .224 \\
\texttt{mfcc2\_sma3\_stddevNorm} & $-$0.01 & [$-$0.05, 0.02] & .520 \\
\texttt{mfcc3\_sma3\_amean} & $-$0.06 & [$-$0.10, $-$0.03] & $<$.001 \\
\texttt{mfcc3\_sma3\_stddevNorm} & 0.00 & [$-$0.03, 0.03] & .974 \\
\texttt{mfcc4\_sma3\_amean} & $-$0.06 & [$-$0.09, $-$0.02] & .002 \\
\texttt{mfcc4\_sma3\_stddevNorm} & $-$0.02 & [$-$0.05, 0.02] & .430 \\
\texttt{logRelF0-H1-H2\_sma3nz\_amean}$^{14}$ & \textbf{$-$0.17} & [$-$0.20, $-$0.13] & $<$.001 \\
\texttt{logRelF0-H1-H2\_sma3nz\_stddevNorm} & 0.02 & [$-$0.02, 0.05] & .457 \\
\texttt{logRelF0-H1-A3\_sma3nz\_amean}$^{14}$ & \textbf{$-$0.13} & [$-$0.16, $-$0.09] & $<$.001 \\
\texttt{logRelF0-H1-A3\_sma3nz\_stddevNorm} & $-$0.01 & [$-$0.04, 0.03] & .705 \\
\texttt{F1amplitudeLogRelF0\_sma3nz\_amean}$^{15}$ & 0.07 & [0.04, 0.11] & $<$.001 \\
\texttt{F1amplitudeLogRelF0\_sma3nz\_stddevNorm} & $-$0.08 & [$-$0.11, $-$0.04] & $<$.001 \\
\texttt{F2amplitudeLogRelF0\_sma3nz\_amean} & 0.09 & [0.05, 0.12] & $<$.001 \\
\texttt{F2amplitudeLogRelF0\_sma3nz\_stddevNorm} & $-$0.10 & [$-$0.13, $-$0.06] & $<$.001 \\
\texttt{F3amplitudeLogRelF0\_sma3nz\_amean} & 0.09 & [0.05, 0.12] & $<$.001 \\
\texttt{F3amplitudeLogRelF0\_sma3nz\_stddevNorm} & \textbf{$-$0.10} & [$-$0.13, $-$0.07] & $<$.001 \\
\texttt{alphaRatioV\_sma3nz\_amean}$^{16}$ & 0.09 & [0.06, 0.13] & $<$.001 \\
\texttt{alphaRatioV\_sma3nz\_stddevNorm} & \textbf{$-$0.12} & [$-$0.16, $-$0.09] & $<$.001 \\
\texttt{hammarbergIndexV\_sma3nz\_amean}$^{17}$ & $-$0.04 & [$-$0.08, $-$0.01] & .021 \\
\texttt{hammarbergIndexV\_sma3nz\_stddevNorm} & 0.06 & [0.03, 0.10] & $<$.001 \\
\texttt{slopeV0-500\_sma3nz\_amean}$^{18}$ & \textbf{0.20} & [0.17, 0.23] & $<$.001 \\
\texttt{slopeV0-500\_sma3nz\_stddevNorm} & 0.00 & [$-$0.03, 0.04] & .919 \\
\texttt{slopeV500-1500\_sma3nz\_amean}$^{19}$ & 0.02 & [$-$0.01, 0.05] & .333 \\
\texttt{slopeV500-1500\_sma3nz\_stddevNorm} & 0.01 & [$-$0.02, 0.04] & .636 \\
\texttt{spectralFluxV\_sma3nz\_amean} & \textbf{0.35} & [0.32, 0.38] & $<$.001 \\
\texttt{spectralFluxV\_sma3nz\_stddevNorm} & \textbf{0.17} & [0.13, 0.20] & $<$.001 \\
\texttt{mfcc1V\_sma3nz\_amean} & \textbf{$-$0.12} & [$-$0.15, $-$0.09] & $<$.001 \\
\texttt{mfcc1V\_sma3nz\_stddevNorm} & \textbf{0.11} & [0.08, 0.15] & $<$.001 \\
\texttt{mfcc2V\_sma3nz\_amean} & $-$0.04 & [$-$0.07, $-$0.01] & .030 \\
\texttt{mfcc2V\_sma3nz\_stddevNorm} & $-$0.01 & [$-$0.04, 0.03] & .845 \\
\texttt{mfcc3V\_sma3nz\_amean} & $-$0.07 & [$-$0.11, $-$0.04] & $<$.001 \\
\texttt{mfcc3V\_sma3nz\_stddevNorm} & 0.02 & [$-$0.01, 0.06] & .224 \\
\texttt{mfcc4V\_sma3nz\_amean} & $-$0.01 & [$-$0.04, 0.02] & .659 \\
\texttt{mfcc4V\_sma3nz\_stddevNorm} & 0.01 & [$-$0.02, 0.05] & .579 \\
\texttt{alphaRatioUV\_sma3nz\_amean} & \textbf{0.13} & [0.10, 0.17] & $<$.001 \\
\texttt{hammarbergIndexUV\_sma3nz\_amean} & \textbf{$-$0.14} & [$-$0.17, $-$0.11] & $<$.001 \\
\texttt{slopeUV0-500\_sma3nz\_amean} & \textbf{0.17} & [0.13, 0.20] & $<$.001 \\
\texttt{slopeUV500-1500\_sma3nz\_amean} & $-$0.05 & [$-$0.08, $-$0.01] & .009 \\
\texttt{spectralFluxUV\_sma3nz\_amean} & \textbf{0.27} & [0.24, 0.30] & $<$.001 \\
\bottomrule
\end{tabular}%
}
\par\vspace{2pt}
\begin{minipage}{\textwidth}\scriptsize
\textit{What a higher value of each acoustic feature means:} $^{1}$\,F0 (in semitones relative to 27.5\,Hz): higher pitch. $^{2}$\,Jitter: more irregular pitch periods (rougher voice). $^{3}$\,Formant frequency: reflects vowel articulation (e.g., higher F1 = more open mouth). $^{4}$\,Formant bandwidth: wider, more damped resonance. $^{5}$\,Loudness: louder perceived voice. $^{6}$\,Shimmer: more irregular amplitude (rougher voice). $^{7}$\,HNR: more periodic, less breathy or hoarse voice. $^{8}$\,Equivalent sound level: higher overall sound level. $^{9}$\,Loudness peaks / s: more syllable-like loudness peaks per second (faster speech). $^{10}$\,Voiced segments / s: more voiced stretches per second. $^{11}$\,Segment length: longer voiced or unvoiced stretches (unvoiced stretches include pauses and voiceless consonants). $^{12}$\,Spectral flux: faster change of the spectrum over time. $^{13}$\,MFCC1: relatively more low-frequency energy; MFCC2--4 describe finer spectral shape and lack a simple interpretation. $^{14}$\,H1--H2, H1--A3: breathier, more relaxed phonation; lower values indicate a tenser, pressed voice. $^{15}$\,Formant relative amplitude: stronger formant relative to the F0 harmonic. $^{16}$\,Alpha ratio: relatively more energy above 1\,kHz. $^{17}$\,Hammarberg index: relatively less energy in 2--5\,kHz. $^{18}$\,Slope 0--500\,Hz: spectrum rises more steeply toward 500\,Hz. $^{19}$\,Slope 500--1500\,Hz: spectrum falls less steeply in this band.
\end{minipage}
\caption{Within-speaker association between all 88 eGeMAPSv02 \citep{7160715} features obtained by openSMILE \citep{eyben2010opensmile} and self-reported pain score on the training speakers, grouped by feature family. Pain and each feature are centered within speaker and normalized; $\beta$ (range $-1$ to $1$) is the linear-regression slope, equivalent to the within-speaker correlation. Positive $\beta$ means pain is higher when the feature is above the speaker's usual value; larger $|\beta|$ means a stronger association. 95\% CI: Wald interval of the fixed effect ($\beta$ ± 1.96 SE); an interval excluding 0 indicates a reliable association. $q$: Benjamini--Hochberg FDR-corrected $p$-value over all 88 features; smaller means less likely to be a false positive. Bold marks features with a significant and non-negligible association ($q<.05$ and $|\beta|\geq .10$). Thresholds are applied before rounding. Feature names follow openSMILE eGeMAPSv02 output.}
\label{tab:egemaps-association}
\end{table*}

Figure~\ref{fig:confusion} shows the corresponding confusion matrices.

\section{Acoustic Correlates of Self-Reported Pain}
\label{app:acoustic-correlates}

\subsection{Feature--Pain Association}
To identify which acoustic properties track pain, we extract all 88 eGeMAPSv02 features \citep{7160715} with openSMILE \citep{eyben2010opensmile} from the recordings of the 41 training speakers. Because both vocal characteristics and pain-rating habits vary across individuals, we normalize the self-reported pain score across all training clips, center and normalize within speaker, and fit a separate linear mixed-effects model for each feature with a random intercept per speaker. Table~\ref{tab:egemaps-association} reports the full results.

\subsection{Feature Groups}

To summarize the significant features at a more interpretable level, we group the 34 features marked in bold in Table~\ref{tab:egemaps-association} into 9 cue groups according to the acoustic property they describe (Table~\ref{tab:cue-groups}). 

\begin{table*}[!ht]
\centering
\scriptsize
\renewcommand{\arraystretch}{1.1}
\begin{tabular*}{\textwidth}{@{\extracolsep{\fill}}ll@{}}
\toprule
\textbf{Group} / eGeMAPSv02 feature & \textbf{Group} / eGeMAPSv02 feature \\
\midrule
\textbf{Pitch level} & \textbf{Formants} \\
\quad \texttt{F0semitoneFrom27.5Hz\_sma3nz\_percentile50.0} & \quad \texttt{F1frequency\_sma3nz\_stddevNorm} \\
\quad \texttt{F0semitoneFrom27.5Hz\_sma3nz\_percentile80.0} & \quad \texttt{F3amplitudeLogRelF0\_sma3nz\_stddevNorm} \\
\textbf{Pitch variability} & \textbf{Spectral flux} \\
\quad \texttt{F0semitoneFrom27.5Hz\_sma3nz\_stddevNorm} & \quad \texttt{spectralFlux\_sma3\_amean} \\
\quad \texttt{F0semitoneFrom27.5Hz\_sma3nz\_pctlrange0-2} & \quad \texttt{spectralFlux\_sma3\_stddevNorm} \\
\textbf{Loudness level} & \quad \texttt{spectralFluxV\_sma3nz\_amean} \\
\quad \texttt{loudness\_sma3\_amean} & \quad \texttt{spectralFluxV\_sma3nz\_stddevNorm} \\
\quad \texttt{loudness\_sma3\_percentile20.0} & \quad \texttt{spectralFluxUV\_sma3nz\_amean} \\
\quad \texttt{loudness\_sma3\_percentile50.0} & \textbf{MFCC1} \\
\quad \texttt{loudness\_sma3\_percentile80.0} & \quad \texttt{mfcc1\_sma3\_amean} \\
\quad \texttt{equivalentSoundLevel\_dBp} & \quad \texttt{mfcc1\_sma3\_stddevNorm} \\
\textbf{Loudness dynamics} & \quad \texttt{mfcc1V\_sma3nz\_amean} \\
\quad \texttt{loudness\_sma3\_stddevNorm} & \quad \texttt{mfcc1V\_sma3nz\_stddevNorm} \\
\quad \texttt{loudness\_sma3\_pctlrange0-2} & \textbf{Spectral tilt} \\
\quad \texttt{loudness\_sma3\_meanRisingSlope} & \quad \texttt{slopeV0-500\_sma3nz\_amean} \\
\quad \texttt{loudness\_sma3\_stddevRisingSlope} & \quad \texttt{slopeUV0-500\_sma3nz\_amean} \\
\quad \texttt{loudness\_sma3\_meanFallingSlope} & \quad \texttt{alphaRatioV\_sma3nz\_stddevNorm} \\
\quad \texttt{loudness\_sma3\_stddevFallingSlope} & \quad \texttt{alphaRatioUV\_sma3nz\_amean} \\
\textbf{Voice quality} & \quad \texttt{hammarbergIndexUV\_sma3nz\_amean} \\
\quad \texttt{HNRdBACF\_sma3nz\_stddevNorm} &  \\
\quad \texttt{logRelF0-H1-H2\_sma3nz\_amean} &  \\
\quad \texttt{logRelF0-H1-A3\_sma3nz\_amean} &  \\
\bottomrule
\end{tabular*}
\caption{Feature groups in Figure~\ref{fig:layerwise}(c). Each line in the figure is the mean held-out $R^2$ over the features listed for its group. The 34 features are those marked in bold in Table~\ref{tab:egemaps-association}.}
\label{tab:cue-groups}
\end{table*}

\section{Perturbation Analysis of the Probing Head}\label{sec:perturb_prob}

To test whether these cues are also used by a model that detects pain, we perturb each test clip by approximately $\pm$1 within-speaker SD along a single acoustic attribute and measure the change in the output of our best binary probe (Table~\ref{tab:probing}), relative to a matched null resynthesis. Table~\ref{tab:perturbation-construction} lists the manipulation for each attribute and verifies that it moves the target attribute while controlling other feature families variation. Jitter, shimmer, and breathiness are excluded because the manipulation often failed to change the target.

Table~\ref{tab:perturbation-encoder1} reports the effect of each perturbation on the probe's pain logit. Spectral balance and pauses have the largest credible effects: more high-frequency energy pushes the probe toward pain, and longer pauses push it away. Loudness and mean pitch have smaller credible effects, despite loudness showing the strongest association with pain in Table~\ref{tab:egemaps-association}. The speech-rate effect is attributable to resynthesis artifacts. 

\begin{table*}[!ht]
\centering\scriptsize
\setlength{\tabcolsep}{3pt}
\resizebox{\textwidth}{!}{%
\begin{tabular}{@{}l l l l r l r r@{}}
\toprule
Cue & Target feature & Manipulation method & Range & Target shift & Largest off-target feature & Off-target shift & Pairs \\
\midrule
Spectral balance & \texttt{alphaRatioV\_sma3nz\_amean} & High-shelf filter at 1\,kHz & $+$4 dB & \textbf{0.89} & \texttt{mfcc1V\_sma3nz\_amean} & 1.46 & 760 \\
 & $-$\texttt{hammarbergIndexV\_sma3nz\_amean} & (SciPy biquad) & $-$4 dB & \textbf{$-$0.87} & \texttt{mfcc1V\_sma3nz\_amean} & 1.36 &  \\
 & \texttt{slopeV0-500\_sma3nz\_amean} & & & & & &  \\
 & \texttt{slopeV500-1500\_sma3nz\_amean} & & & & & &  \\
\midrule
Loudness & \texttt{loudness\_sma3\_amean} & Waveform gain & $+$1 dB & \textbf{0.71} & \texttt{spectralFluxUV\_sma3nz\_amean} & 1.23 & 760 \\
 & & (NumPy) & $-$2 dB & \textbf{$-$1.28} & \texttt{spectralFluxUV\_sma3nz\_amean} & 2.10 &  \\
\addlinespace
Loudness variability & \texttt{loudness\_sma3\_stddevNorm} & Envelope exponent$^{a}$ & 1.10 & \textbf{0.77} & \texttt{spectralFluxUV\_sma3nz\_amean} & 0.95 & 760 \\
 & & (SciPy Hilbert) & 0.90 & \textbf{$-$0.73} & \texttt{spectralFluxUV\_sma3nz\_amean} & 0.97 &  \\
\midrule
Mean pitch & \texttt{F0semitoneFrom27.5Hz\_sma3nz\_amean} & PSOLA F0 shift & $+$1 st & \textbf{1.00} & \texttt{jitterLocal\_sma3nz\_stddevNorm} & 0.79 & 760 \\
 & & (Praat) & $-$1 st & \textbf{$-$0.98} & \texttt{jitterLocal\_sma3nz\_stddevNorm} & 0.82 &  \\
\addlinespace
Pitch variability & \texttt{F0semitoneFrom27.5Hz\_sma3nz\_stddevNorm} & PSOLA F0-range scaling & $\times$1.75 & \textbf{1.20} & \texttt{F3bandwidth\_sma3nz\_stddevNorm} & 1.02 & 760 \\
 & \texttt{F0semitoneFrom27.5Hz\_sma3nz\_pctlrange0-2} & (Praat) & $\times$0.25 & \textbf{$-$1.16} & \texttt{F1bandwidth\_sma3nz\_stddevNorm} & 0.77 &  \\
\midrule
Speech rate & \texttt{VoicedSegmentsPerSec} & Phase-vocoder stretch & $\times$1.35 & \textbf{0.78} & \texttt{loudness\_sma3\_percentile20.0} & 2.53 & 760 \\
 & $-$\texttt{MeanVoicedSegmentLengthSec} & (librosa) & $\times$0.75 & \textbf{$-$1.04} & \texttt{loudness\_sma3\_meanFallingSlope} & 2.22 &  \\
\addlinespace
Pauses & \texttt{MeanUnvoicedSegmentLength} & Non-speech resampling$^{b}$ & $\times$1.75 & \textbf{0.77} & \texttt{slopeUV0-500\_sma3nz\_amean} & 4.12 & 755$^{d}$ \\
 & & (WebRTC VAD) & $\times$0.25 & \textbf{$-$0.76} & \texttt{loudness\_sma3\_percentile20.0} & 3.55 &  \\
\midrule
Jitter$^{\dagger}$ & \texttt{jitterLocal\_sma3nz\_amean} & F0 frame noise & $\sigma$ 0.02 & \textbf{0.54} & \texttt{F1bandwidth\_sma3nz\_amean} & 0.84 & 760 \\
 & & (WORLD) & smooth 0.04 & $-$0.18$^{c}$ & \texttt{F1bandwidth\_sma3nz\_stddevNorm} & 0.85 & -- \\
\addlinespace
Shimmer$^{\dagger}$ & \texttt{shimmerLocaldB\_sma3nz\_amean} & Frame-gain noise & $\sigma$ 0.10 & \textbf{1.12} & \texttt{F3bandwidth\_sma3nz\_stddevNorm} & 0.29 & 760 \\
 & & (WORLD) & smoothed $\sigma$ 0.02 & $-$0.01 & \texttt{F3bandwidth\_sma3nz\_amean} & 0.06 & -- \\
\addlinespace
Breathiness$^{\dagger}$ & $-$\texttt{HNRdBACF\_sma3nz\_amean} & Aperiodicity scaling & mix 0.20 & \textbf{1.28} & \texttt{F3bandwidth\_sma3nz\_stddevNorm} & 0.68 & 760 \\
 & & (WORLD) & reduce 0.50 & $-$0.02 & \texttt{F3bandwidth\_sma3nz\_stddevNorm} & 0.42 & -- \\
\bottomrule
\end{tabular}%
}
\caption{Construction and validation of the audio perturbations. We select ten representative acoustic attributes spanning the four parameter groups across 88 features. Each manipulation aims to shift its target eGeMAPSv02 parameter(s) while leaving parameters of other feature families unchanged (parameters within one family are inherently correlated). Shifts are measured on 24 recordings, each from a different \emph{training} speaker, as the perturbed recording minus its matching null resynthesis (same tool at zero strength), divided by that parameter's mean within-speaker SD over the training speakers. \textbf{Target shift} is the median of this shift over the 24 recordings, oriented so that a leading $-$ on a target parameter means it is expected to decrease, and averaged over the target parameters when a cue has several. \textbf{Range}: for each cue and direction, the candidate strength whose Target shift is closest to $\pm$1\,SD. \textbf{Largest off-target feature}: the parameter with the largest median absolute shift after excluding the target's own family (e.g., all F0 parameters for the pitch cues). Its ratio to $|$Target shift$|$ is flagged when above 0.5 but is only reported, not used for filtering. RMS level is matched to the original recording for every manipulation except Loudness. \textbf{Filtering.} A setting passes calibration if $|$Target shift$|\in[0.5, 1.5]$ and none of the 24 recordings fails processing; passing settings are shown in bold, and non-bold settings are excluded. WORLD could not lower jitter, shimmer or breathiness (Target shift $\approx$\,0), so these cues were tested in the increase direction only, contrasting a 2$\times$ against the 1$\times$ strength. \textbf{Pairs}: test recordings (760 in total, from 10 held-out speakers) for which both perturbed versions of a cue were processed successfully; the effect of each bidirectional cue is estimated from the $+$ versus $-$ version of the same recording. Cues marked $^{\dagger}$ were excluded from the analysis because more than 5\% of their 1$\times$/2$\times$ pairs showed no achieved change in the target parameter (jitter 10.0\%, shimmer 8.9\%, breathiness 9.3\%). In total we generated 18{,}240 test-set versions for the 17 passing settings and 7 null resyntheses (4 failed, all Pauses), 2{,}280 2$\times$ WORLD versions, and 4{,}560 round-trip null resyntheses for the PSOLA and speech-rate cues. $^{a}$\,Each sample is multiplied by $(e/\tilde{e})^{k-1}$, where $e$ is the smoothed Hilbert envelope, $\tilde{e}$ its median and $k$ the exponent; $k>1$ exaggerates and $k<1$ flattens loudness fluctuations. $^{b}$\,Non-speech stretches detected by WebRTC VAD are resampled to the given multiple of their length; speech is left unchanged. $^{c}$\,One of the 24 calibration recordings also failed processing. $^{d}$\,760 minus 4 recordings whose perturbation failed and 1 recording whose $+$ and $-$ versions had identical target values.}
\label{tab:perturbation-construction}
\end{table*}

\begin{table*}[!ht]
\centering\small
\begin{tabular*}{\textwidth}{@{\extracolsep{\fill}}l r c r c l c@{}}
\toprule
Attribute & Effect & 95\% CI & Sym. & Null slopes & Verdict & Flip (+ / $-$) \\
\midrule
Spectral balance & 0.62 & [0.59, 0.66] & 0.05 & 0.62 / -- & credible & 0.22 / 0.18 \\
Loudness & 0.13 & [0.11, 0.14] & $-$0.03 & 0.13 / -- & credible & 0.03 / 0.04 \\
Loudness variability & 0.02 & [0.01, 0.05] & 0.01 & 0.02 / -- & credible & 0.01 / 0.01 \\
Mean pitch & 0.19 & [0.04, 0.34] & $-$0.20 & 0.10 / 0.24 & credible & 0.11 / 0.18 \\
Pitch variability & $-$0.23 & [$-$0.43, $-$0.001] & $-$0.23 & 0.06 / $-$0.17 & no effect & 0.19 / 0.12 \\
Speech rate & $-$0.14 & [$-$1.03, 0.23] & 0.57 & $-$1.84 / 0.06 & artifact & 0.22 / 0.26 \\
Pauses$^\ddagger$ & $-$0.68 & [$-$1.44, $-$0.11] & 0.07 & -- / -- & credible & 0.31 / 0.24 \\
\bottomrule
\end{tabular*}
\caption{Effect of each perturbation on the binary pain logit of the CRNN probe at audio-encoder block~1 (760 held-out clips, 10 speakers; perturbations as in Table~\ref{tab:perturbation-construction}; $^\ddagger$755 after one zero-dose pair). Effect: $(y^{+}-y^{-})/(s^{+}-s^{-})$ averaged over clips and three seeds, where $y$ is the pain logit in units of its SD on original test clips and $s$ the achieved target shift in within-speaker SD; positive means raising the attribute pushes toward \emph{pain}. CI: speaker bootstrap. Sym.: $(y^{+}+y^{-})/2-y^{0}$, the direction-independent change. Null slopes (standard / round-trip): one-sided slopes against the standard and round-trip null resynthesis (round-trip only for PSOLA and time-stretch; undefined for pauses because some clips had zero shift relative to the null). Credible: effect CI excludes 0 and the two null slopes agree in sign; artifact: effect CI includes 0 but the Sym.\ CI excludes 0. Flip: fraction of binary predictions that change relative to the original audio under the + and $-$ perturbation.}
\label{tab:perturbation-encoder1}
\end{table*}

\section{Acoustic Perception Dataset Creation}
\label{app:contrast-dataset}

\paragraph{Stimuli.}
We sample 100 neutral Harvard-sentence recordings from TAME-Pain and render each twice, at a low and a high magnitude of a single acoustic cue, yielding 100 low/high pairs per cue (Table~\ref{tab:max-contrast}). We initially manipulated six cues, but excluded voice quality after a pilot listening check showed that its contrast was not reliably distinguishable, leaving five cues for all analyses.

Magnitudes are fixed per cue rather than tuned per recording, and are set far beyond natural within-speaker variation so that the contrast is unambiguous to human listeners. All clips except those for the loudness cue are RMS-matched to their source recording. Since the goal is an obvious contrast on the target cue, we do not otherwise control side effects on non-target acoustic qualities.

Pitch manipulations use PSOLA resynthesis~\citep{moulines1990psola} in Praat~\citep{boersma2024praat} via Parselmouth~0.4.6~\citep{Parselmouth}. Acoustic features are eGeMAPSv02 functionals extracted with openSMILE~2.5.1~\citep{eyben2010opensmile}.

\begin{table*}[!ht]
\centering
\resizebox{\textwidth}{!}{%
\begin{tabular}{@{}l l l r r r r@{}}
\toprule
Cue & Target feature(s) & Manipulation method & Low & High & $\Delta$ target & Pairs \\
\midrule
Pitch level & \texttt{F0semitoneFrom27.5Hz\_sma3nz\_amean} & PSOLA F0 shift (Praat) & $-$8 st & $+$8 st & \textbf{15.87} & 100 \\
\addlinespace
Pitch variability & \texttt{F0semitoneFrom27.5Hz\_sma3nz\_stddevNorm} & PSOLA F0-range scaling (Praat) & $\times$0.0 & $\times$2.5 & \textbf{3.96} & 100 \\
 & \texttt{F0semitoneFrom27.5Hz\_sma3nz\_pctlrange0-2} & & & & & \\
\midrule
Loudness level & \texttt{loudness\_sma3\_amean} & Waveform gain & $-$20 dB & $+$20 dB & \textbf{30.95} & 100 \\
 & \texttt{equivalentSoundLevel\_dBp} & & & & & \\
\addlinespace
Loudness dynamics & \texttt{loudness\_sma3\_stddevNorm} & Envelope exponent (Hilbert) & 0.3 & 2.0 & \textbf{10.62} & 100 \\
\midrule
Spectral tilt & \texttt{alphaRatioV\_sma3nz\_amean} & High-shelf filter at 1\,kHz & $-$15 dB & $+$15 dB & \textbf{11.88} & 100 \\
 & $-$\texttt{hammarbergIndexV\_sma3nz\_amean} & & & & & \\
 & \texttt{slopeV0-500\_sma3nz\_amean} & & & & & \\
 & \texttt{slopeV500-1500\_sma3nz\_amean} & & & & & \\
\midrule
Voice quality$^{\dagger}$ & $-$\texttt{logRelF0-H1-H2\_sma3nz\_amean} & Harmonic gain, H1 $+g$ / H2 $-g$ (STFT) & $-$15 dB & $+$15 dB & 5.15 & 100 \\
\bottomrule
\end{tabular}%
}
\caption{Construction of the maximal low/high acoustic contrasts. Each of 100
neutral Harvard-sentence recordings from TAME-Pain is rendered once at the low
and once at the high magnitude. $\Delta$ target is the median high-minus-low
difference in the target composite (in within-speaker SD units); every pair
moved in the intended direction (100/100 per analysed cue).
$^{\dagger}$Generated but excluded from analysis.}
\label{tab:max-contrast}
\end{table*}

\paragraph{Human validation.}
To verify that the contrasts are perceptible, three listeners (one author and two volunteers) annotated a random subset of 100 pairs (20 per cue), using the same question as the models. Table~\ref{tab:human-listeners} reports per-listener accuracy.
The two non-author listeners participated voluntarily without payment after giving informed consent. The task was short and collected no personal data.

\begin{table}[!ht]
\centering
\small
\begin{tabular}{lcccc}
\toprule
Feature & L1 & L2 & L3 & Pooled \\
\midrule
Pitch level        & 100 & 100 & 100 & 100.0 \\
Loudness level     & 100 & 100 & 100 & 100.0 \\
Spectral tilt      & 100 & 100 & 100 & 100.0 \\
Pitch variability  & 100 &  90 & 100 &  96.7 \\
Loudness dynamics  & 100 &  95 &  95 &  96.7 \\
\midrule
Overall            & 100 &  97 &  99 &  98.7 \\
\bottomrule
\end{tabular}
\caption{Per-listener accuracy (\%) on the acoustic perception test
($n=20$ pairs per feature). Fleiss' $\kappa=0.95$ across the three listeners;
all three agreed on 96 of 100 pairs, and the majority vote was correct on all 100.}
\label{tab:human-listeners}
\end{table}

\section{Prompt with Cues Additional Results}
\label{app:cue-additional}

Table~\ref{tab:parsing-acoustic} reports parsing success rates for the cue-augmented prompts; all models except gemini-3.5-flash parse nearly all outputs, while gemini-3.5-flash fails more often in the 20-shot settings (83.0--92.7\%), partly explaining its low 20-shot scores. Table~\ref{tab:acoustic-cue-delta-zeroshot} reports the zero-shot single-recording setting, where adding cues yields no meaningful gain on neutral sentences, and AF alone stays at chance on both recording types. On spoken pain ratings, adding cues slightly lowers three-class accuracy for gpt-realtime-mini and Qwen2.5-Omni (by up to 5.3 points), suggesting that the cues can distract from the lexical content.

\begin{table*}[!ht]
\centering
\resizebox{\textwidth}{!}{%
\begin{tabular}{l cc cc cc cc cc cc cc}
\toprule
& \multicolumn{2}{c}{gpt-realtime-mini}
& \multicolumn{2}{c}{gpt-realtime}
& \multicolumn{2}{c}{gemini-3.5-flash}
& \multicolumn{2}{c}{gemini-3.1-pro-preview}
& \multicolumn{2}{c}{Qwen2.5-Omni-7B}
& \multicolumn{2}{c}{Qwen3-Omni-30B-A3B}
& \multicolumn{2}{c}{Kimi-Audio} \\
\cmidrule(lr){2-3}\cmidrule(lr){4-5}\cmidrule(lr){6-7}\cmidrule(lr){8-9}\cmidrule(lr){10-11}\cmidrule(lr){12-13}\cmidrule(lr){14-15}
Modality & Binary & 3-class & Binary & 3-class & Binary & 3-class & Binary & 3-class
         & Binary & 3-class & Binary & 3-class & Binary & 3-class \\
\midrule
\multicolumn{15}{l}{\textit{Neutral + Pairwise}} \\
Audio      & 100.0 & 100.0 & 100.0 & 100.0 & 100.0 & 100.0 & 100.0 & 100.0 & 100.0 & 100.0 & 100.0 & 100.0 & 100.0 & 100.0 \\
Text+Audio & 100.0 & 100.0 & 100.0 & 100.0 & 99.5  & 99.6  & 100.0 & 100.0 & 100.0 & 100.0 & 100.0 & 100.0 & 100.0 & 100.0 \\
AF         & 100.0 & 100.0 & 100.0 & 100.0 & 100.0 & 98.9  & 100.0 & 100.0 & 100.0 & 100.0 & 100.0 & 100.0 & --    & --    \\
\midrule
\multicolumn{15}{l}{\textit{Neutral + 20-shot}} \\
Audio      & 100.0 & 100.0 & 100.0 & 100.0 & 87.3  & 83.0  & 100.0 & 100.0 & 100.0 & 100.0 & 100.0 & 100.0 & 100.0 & 100.0 \\
Text+Audio & 100.0 & 100.0 & 100.0 & 100.0 & 90.7  & 85.0  & 99.7  & 100.0 & 100.0 & 100.0 & 100.0 & 100.0 & 100.0 & 100.0 \\
AF         & 100.0 & 99.5  & 100.0 & 100.0 & 92.7  & 91.5  & 99.7  & 99.5  & 100.0 & 100.0 & 100.0 & 100.0 & --    & --    \\
\midrule
Overall    & 100.0 & 99.9  & 100.0 & 100.0 & 94.2  & 93.9  & 99.9  & 99.9  & 100.0 & 100.0 & 100.0 & 100.0 & 100.0 & 100.0 \\
\bottomrule
\end{tabular}%
}
\caption{Parsing success rate (\%) of model outputs when acoustic cues are added to the prompt (settings as in Table~\ref{tab:acoustic-cue-delta-neutral}). A response counts as successfully parsed if it contains exactly one valid label. Unparsed responses are counted as incorrect. -- indicates not evaluated.}
\label{tab:parsing-acoustic}
\end{table*}

\begin{table*}[!ht]
\centering
\setlength{\tabcolsep}{3pt}
\scalebox{\tabscale}{%
\begin{tabular}{l ccc ccc ccc ccc}
\toprule
& \multicolumn{6}{c}{\textbf{Pain vs.\ No Pain}}
& \multicolumn{6}{c}{\textbf{3-Class Pain Severity}} \\
\cmidrule(lr){2-7}\cmidrule(lr){8-13}
& \multicolumn{3}{c}{Neutral Harvard Sentence} & \multicolumn{3}{c}{Spoken Pain Rating}
& \multicolumn{3}{c}{Neutral Harvard Sentence} & \multicolumn{3}{c}{Spoken Pain Rating} \\
\cmidrule(lr){2-4}\cmidrule(lr){5-7}\cmidrule(lr){8-10}\cmidrule(lr){11-13}
Model & Audio & Text+Audio & AF & Audio & Text+Audio & AF & Audio & Text+Audio & AF & Audio & Text+Audio & AF \\
\midrule
gpt-realtime-mini & \textbf{50.8}\,\dgain{2.7} & 50.3\,\dgain{0.1} & 49.8 & 98.5\,\ddrop{0.2} & 98.2\,\ddrop{0.5} & \textbf{50.5} & \textbf{34.0}\,\dgain{0.3} & \textbf{33.7}\,\dgain{0.2} & 34.3 & 90.7\,\ddrop{5.3} & 88.3\,\ddrop{5.3} & 33.0 \\
gpt-realtime      & 50.0\,\dgain{0.4} & \textbf{51.7}\,\dgain{1.0} & 50.0 & 98.8\,\ddrop{0.1} & 98.9\,\dsame      & 50.0 & 31.2\,\ddrop{1.6} & 33.4\,\ddrop{0.7} & 33.3 & 98.9\,\dsame      & 98.8\,\ddrop{0.5} & \textbf{34.3} \\
\midrule
Qwen2.5-Omni-7B   & 50.0\,\dgain{0.2} & 50.1\,\dgain{0.1} & \textbf{50.4} & 98.1\,\ddrop{0.7} & 98.8\,\dgain{0.2} & 48.4 & 33.4\,\dsame      & 33.4\,\dgain{0.3} & 33.9 & 93.6\,\ddrop{4.4} & 89.6\,\ddrop{3.7} & 33.4 \\
Qwen3-Omni-30B    & 50.3\,\dgain{0.1} & 49.9\,\ddrop{0.2} & 50.2 & \textbf{99.5}\,\dsame      & \textbf{99.2}\,\dgain{0.1} & 50.1 & 33.4\,\dsame      & 33.3\,\dsame      & \textbf{34.5} & \textbf{99.5}\,\dsame      & \textbf{99.0}\,\dgain{1.2} & 32.1 \\
\bottomrule
\end{tabular}%
}
\caption{Effect of adding acoustic cues to the prompt in the zero-shot single-recording setting (balanced accuracy, \%). We evaluate one closed-source model family (gpt-realtime, top) and one open-weight model family (Qwen-Omni, bottom) due to resource limitation. Audio and Text+Audio cells report results with acoustic cues; arrows give the change relative to the same model and modality without cues. AF (acoustic features only, no audio input) has no cue-free counterpart and is reported as-is. Best result per column in \textbf{bold}.}
\label{tab:acoustic-cue-delta-zeroshot}
\end{table*}

\begin{figure*}[t]
  \centering
  \includegraphics[width=\textwidth]{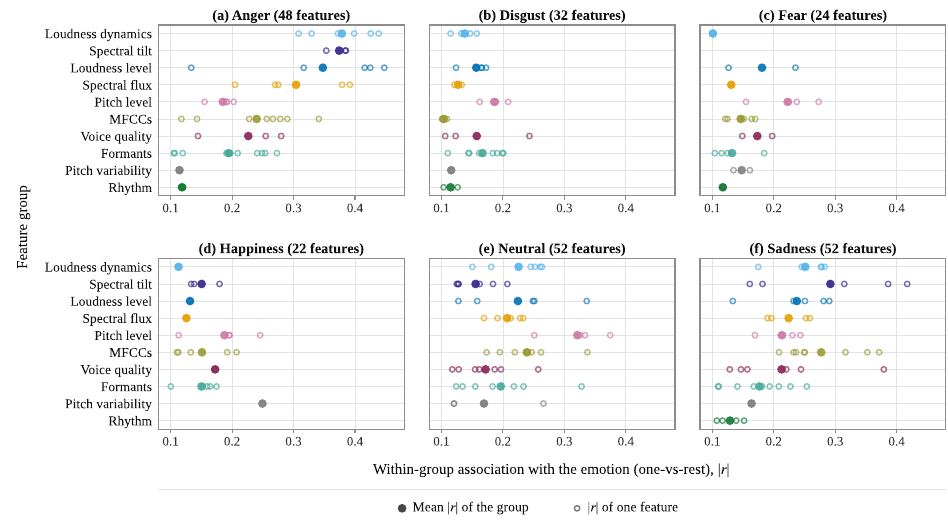}
  \caption{Within-group association between each emotion (one-vs-rest) and each acoustic
  feature group on the LISTEN neutral-text set. Open circles show $|r|$ of individual features
  (within-group point-biserial $r$); filled dots show the group mean. Only features with
  $q < .05$ and $|r| \geq .10$ for that emotion are shown; empty rows mean no feature in the
  group passed. Rows follow the same order in every panel (mean $|r|$ across emotions).
  Surprise is omitted because too few groups contain within-group variation for inference.
  Feature groups are listed in Table~\ref{tab:cue-groups-emotion}.}
  \label{fig:emotion-cue-groups-listen}
\end{figure*}

We further validate the effectiveness of acoustic cues on LISTEN~\citep{chen2026listen} dataset under pairwise comparison against a randomly selected neutral reference as shown in Table~\ref{tab:listen-pairwise}, adding raw acoustic cues brings accuracy gains across models in most of the settings, particularly under the multimodal Text+Audio condition. The prompts we used are in Appendix~\ref{app:prompts-listen-pairwise-acoustic}. The acoustic features are extracted and analyzed using the same method as our main experiment in \S~\ref{sec:rq4}. Our visualized per-emotion type acoustic feature importance is shown in Figure~\ref{fig:emotion-cue-groups-listen}.

For context, \citet{chen2026listen} report 34.0\% accuracy for
Qwen2.5 in the single-utterance audio-only setting, compared with 28.18\% in our reference-conditioned audio setting. Unlike our binary and three-class pain tasks, LISTEN requires
selection among 5--8 candidate labels. One possible explanation is that the larger label space makes it harder for the model to effectively use an additional audio reference.

\begin{table*}[!ht]
\centering
\scriptsize
\renewcommand{\arraystretch}{1.05}
\begin{tabular*}{\textwidth}{@{\extracolsep{\fill}}ll@{}}
\toprule
\textbf{Group} / eGeMAPSv02 feature & \textbf{Group} / eGeMAPSv02 feature \\
\midrule
\textbf{Pitch level} (4) & \textbf{Formants} (14) \\
\quad \texttt{F0semitoneFrom27.5Hz\_sma3nz\_amean}$^\dagger$ & \quad \texttt{F1frequency\_sma3nz\_amean}$^\dagger$ \\
\quad \texttt{F0semitoneFrom27.5Hz\_sma3nz\_percentile20.0} & \quad \texttt{F1frequency\_sma3nz\_stddevNorm} \\
\quad \texttt{F0semitoneFrom27.5Hz\_sma3nz\_percentile50.0} & \quad \texttt{F1bandwidth\_sma3nz\_amean} \\
\quad \texttt{F0semitoneFrom27.5Hz\_sma3nz\_percentile80.0} & \quad \texttt{F1bandwidth\_sma3nz\_stddevNorm} \\
\textbf{Pitch variability} (5) & \quad \texttt{F1amplitudeLogRelF0\_sma3nz\_amean} \\
\quad \texttt{F0semitoneFrom27.5Hz\_sma3nz\_stddevNorm} & \quad \texttt{F1amplitudeLogRelF0\_sma3nz\_stddevNorm} \\
\quad \texttt{F0semitoneFrom27.5Hz\_sma3nz\_pctlrange0-2}$^\dagger$ & \quad \texttt{F2frequency\_sma3nz\_amean} \\
\quad \texttt{F0semitoneFrom27.5Hz\_sma3nz\_meanFallingSlope} & \quad \texttt{F2bandwidth\_sma3nz\_stddevNorm} \\
\quad \texttt{F0semitoneFrom27.5Hz\_sma3nz\_stddevFallingSlope} & \quad \texttt{F2amplitudeLogRelF0\_sma3nz\_amean} \\
\quad \texttt{F0semitoneFrom27.5Hz\_sma3nz\_stddevRisingSlope} & \quad \texttt{F2amplitudeLogRelF0\_sma3nz\_stddevNorm} \\
\textbf{Loudness level} (5) & \quad \texttt{F3frequency\_sma3nz\_amean} \\
\quad \texttt{loudness\_sma3\_amean} & \quad \texttt{F3bandwidth\_sma3nz\_amean} \\
\quad \texttt{loudness\_sma3\_percentile20.0} & \quad \texttt{F3amplitudeLogRelF0\_sma3nz\_amean} \\
\quad \texttt{loudness\_sma3\_percentile50.0} & \quad \texttt{F3amplitudeLogRelF0\_sma3nz\_stddevNorm} \\
\quad \texttt{loudness\_sma3\_percentile80.0} & \textbf{Spectral flux} (5) \\
\quad \texttt{equivalentSoundLevel\_dBp}$^\dagger$ & \quad \texttt{spectralFlux\_sma3\_amean}$^\dagger$ \\
\textbf{Loudness dynamics} (6) & \quad \texttt{spectralFlux\_sma3\_stddevNorm} \\
\quad \texttt{loudness\_sma3\_stddevNorm} & \quad \texttt{spectralFluxV\_sma3nz\_amean} \\
\quad \texttt{loudness\_sma3\_pctlrange0-2} & \quad \texttt{spectralFluxV\_sma3nz\_stddevNorm} \\
\quad \texttt{loudness\_sma3\_meanRisingSlope}$^\dagger$ & \quad \texttt{spectralFluxUV\_sma3nz\_amean} \\
\quad \texttt{loudness\_sma3\_stddevRisingSlope} & \textbf{Spectral tilt} (7) \\
\quad \texttt{loudness\_sma3\_meanFallingSlope} & \quad \texttt{alphaRatioV\_sma3nz\_amean}$^\dagger$ \\
\quad \texttt{loudness\_sma3\_stddevFallingSlope} & \quad \texttt{alphaRatioUV\_sma3nz\_amean} \\
\textbf{Rhythm} (5) & \quad \texttt{hammarbergIndexV\_sma3nz\_amean} \\
\quad \texttt{MeanUnvoicedSegmentLength}$^\dagger$ & \quad \texttt{hammarbergIndexV\_sma3nz\_stddevNorm} \\
\quad \texttt{StddevUnvoicedSegmentLength} & \quad \texttt{hammarbergIndexUV\_sma3nz\_amean} \\
\quad \texttt{StddevVoicedSegmentLengthSec} & \quad \texttt{slopeV0-500\_sma3nz\_amean} \\
\quad \texttt{VoicedSegmentsPerSec} & \quad \texttt{slopeUV500-1500\_sma3nz\_amean} \\
\quad \texttt{loudnessPeaksPerSec} & \textbf{MFCCs} (8) \\
\textbf{Voice quality} (8) & \quad \texttt{mfcc1\_sma3\_amean} \\
\quad \texttt{HNRdBACF\_sma3nz\_amean} & \quad \texttt{mfcc1V\_sma3nz\_amean}$^\dagger$ \\
\quad \texttt{logRelF0-H1-H2\_sma3nz\_amean} & \quad \texttt{mfcc2\_sma3\_amean} \\
\quad \texttt{logRelF0-H1-A3\_sma3nz\_amean}$^\dagger$ & \quad \texttt{mfcc2V\_sma3nz\_amean} \\
\quad \texttt{logRelF0-H1-A3\_sma3nz\_stddevNorm} & \quad \texttt{mfcc3\_sma3\_amean} \\
\quad \texttt{jitterLocal\_sma3nz\_amean} & \quad \texttt{mfcc3V\_sma3nz\_amean} \\
\quad \texttt{jitterLocal\_sma3nz\_stddevNorm} & \quad \texttt{mfcc4\_sma3\_amean} \\
\quad \texttt{shimmerLocaldB\_sma3nz\_amean} & \quad \texttt{mfcc4V\_sma3nz\_amean} \\
\quad \texttt{shimmerLocaldB\_sma3nz\_stddevNorm} &  \\
\bottomrule
\end{tabular*}
\caption{Acoustic feature groups for the emotion analysis on the LISTEN neutral-text set, used in Figure~\ref{fig:emotion-cue-groups-listen}. The 67 features are those with $q < .05$ and $|r| \geq .10$ (one-vs-rest, within-group point-biserial $r$) for at least one emotion; numbers in parentheses give the number of features per group. $^\dagger$ marks the feature with the largest mean $|r|$ across emotions in each group.}
\label{tab:cue-groups-emotion}
\end{table*}

\begin{table*}[!ht]
\centering
\setlength{\tabcolsep}{6pt}
\scalebox{\tabscale}{%
\begin{tabular}{l ccc ccc}
\toprule
& \multicolumn{3}{c}{\textbf{Neutral Audio (Pairwise)}}
& \multicolumn{3}{c}{\textbf{Neutral Audio+Text (Pairwise)}} \\
\cmidrule(lr){2-4}\cmidrule(lr){5-7}
Model & Base & +Features & $\Delta$ & Base & +Features & $\Delta$ \\
\midrule
gpt-realtime-mini & 18.38 & 18.41 & \dgain{0.03} & 17.80 & 19.79 & \dgain{1.99} \\
gpt-realtime      & 20.75 & 23.90 & \dgain{3.15} & 18.83 & 22.37 & \dgain{3.54} \\
\midrule
Qwen2.5           & 28.18 & 27.07 & \ddrop{1.11} & 18.47 & 25.99 & \dgain{7.53} \\
Qwen3             & \textbf{28.79} & \textbf{35.94} & \textbf{\dgain{7.15}} & \textbf{25.93} & \textbf{33.72} & \textbf{\dgain{7.79}} \\
\bottomrule
\end{tabular}%
}
\caption{\textbf{Effect of adding raw acoustic cues on LISTEN dataset in pairwise settings} (accuracy, \%). $\Delta$ denotes the change relative to baseline (\dgain{} for improvements, \ddrop{} for drops). Best results and highest gains per condition are in \textbf{bold}.}
\label{tab:listen-pairwise}
\end{table*}

\section{Prompt Templates}
\label{app:prompts}
 
This section lists the prompts used for every setting in Table~\ref{tab:main}, Table~\ref{tab:acoustic-cue-delta-neutral} and Table~\ref{tab:acoustic-cue-delta-zeroshot}. All prompts are multiple-choice and ask the model to
answer with a single option letter. Under the same dataset subset setting, each Text+Audio prompt uses the same items as the corresponding Audio prompt, with transcriptions added. All choices are randomly shuffled to reduce the effect of position bias.

In all prompts, we describe the pain rating scale as 0--10 instead of the defined 1--10, because some speakers report a pain level of 0 even when asked to rate on a 1--10 scale. In the prompts,
\textcolor{teal!80!black}{\texttt{[AUDIO: \dots]}} marks the position of an
audio input and \textcolor{blue!70!black}{\texttt{\{\dots\}}} marks a text
placeholder filled at request time. Table~\ref{tab:tasks} lists the prompt used for each task.
 
\subsection{Zero-shot Prompts}
\label{app:prompts-zeroshot}
 
These prompts are used for the Neutral Harvard Sentence and Spoken Pain Rating
columns. The model receives only the target recording, its transcription, or
both.
 
\begin{promptbox}{Pain vs.\ No Pain, Text}{prompt:bin-zs-text}
Read the transcription and judge whether the speaker is in pain.
 
Pain is self-reported by the speaker on a scale from 0 to 10. For this task, ratings 0-3 are classified as "No pain" and ratings 4-10 as "Pain".
 
Transcription:\newline
\ph{\{target\_\allowbreak{}transcript\}}
 
Based on the text expression, which is more likely for the speaker in the transcription?\newline
A. Pain\newline
B. No pain
 
Answer with a single letter (A or B) and nothing else. If uncertain, choose the more likely option.
\end{promptbox}
 
\begin{promptbox}{Pain vs.\ No Pain, Audio}{prompt:bin-zs-audio}
Listen to the target recording and judge whether the speaker is in pain.
 
Pain is self-reported by the speaker on a scale from 0 to 10. For this task, ratings 0-3 are classified as "No pain" and ratings 4-10 as "Pain".
 
Target:\newline
\aud{[AUDIO: target\_\allowbreak{}audio]}
 
Based on the vocal expression, which is more likely for the speaker in the target recording?\newline
A. Pain\newline
B. No pain
 
Answer with a single letter (A or B) and nothing else. If uncertain, choose the more likely option.
\end{promptbox}
 
\begin{promptbox}{Pain vs.\ No Pain, Text+Audio}{prompt:bin-zs-ta}
Listen to the target recording and read its transcription to judge whether the speaker is in pain.
 
Pain is self-reported by the speaker on a scale from 0 to 10. For this task, ratings 0-3 are classified as "No pain" and ratings 4-10 as "Pain".
 
The transcription may not fully match the recording verbatim.
 
Target recording:\newline
\aud{[AUDIO: target\_\allowbreak{}audio]}
 
Transcription:\newline
\ph{\{target\_\allowbreak{}transcript\}}
 
Based on the vocal and text expression, which is more likely for the speaker in the target recording and transcription?\newline
A. Pain\newline
B. No pain
 
Answer with a single letter (A or B) and nothing else. If uncertain, choose the more likely option.
\end{promptbox}
 
\begin{promptbox}{3-Class, Text}{prompt:tri-zs-text}
Read the transcription and judge the speaker's pain severity.
 
Pain is self-reported by the speaker on a scale from 0 to 10. For this task, ratings 0-3 are classified as "Mild", 4-6 as "Moderate", and 7-10 as "Severe".
 
Transcription:\newline
\ph{\{target\_\allowbreak{}transcript\}}
 
Based on the text expression, which is more likely for the speaker in the transcription?\newline
A. Mild\newline
B. Moderate\newline
C. Severe
 
Answer with a single letter (A, B, or C) and nothing else. If uncertain, choose the more likely option.
\end{promptbox}
 
\begin{promptbox}{3-Class, Audio}{prompt:tri-zs-audio}
Listen to the target recording and judge the speaker's pain severity.
 
Pain is self-reported by the speaker on a scale from 0 to 10. For this task, ratings 0-3 are classified as "Mild", 4-6 as "Moderate", and 7-10 as "Severe".
 
Target:\newline
\aud{[AUDIO: target\_\allowbreak{}audio]}
 
Based on the vocal expression, which is more likely for the speaker in the target recording?\newline
A. Mild\newline
B. Moderate\newline
C. Severe
 
Answer with a single letter (A, B, or C) and nothing else. If uncertain, choose the more likely option.
\end{promptbox}
 
\begin{promptbox}{3-Class, Text+Audio}{prompt:tri-zs-ta}
Listen to the target recording and read its transcription to judge the speaker's pain severity.
 
Pain is self-reported by the speaker on a scale from 0 to 10. For this task, ratings 0-3 are classified as "Mild", 4-6 as "Moderate", and 7-10 as "Severe".
 
The transcription may not fully match the recording verbatim.
 
Target recording:\newline
\aud{[AUDIO: target\_\allowbreak{}audio]}
 
Transcription:\newline
\ph{\{target\_\allowbreak{}transcript\}}
 
Based on the vocal and text expression, which is more likely for the speaker in the target recording and transcription?\newline
A. Mild\newline
B. Moderate\newline
C. Severe
 
Answer with a single letter (A, B, or C) and nothing else. If uncertain, choose the more likely option.
\end{promptbox}
 
\subsection{Pairwise Prompts}
\label{app:prompts-pairwise}
 
These prompts are used for the Neutral + Pairwise columns. Both recordings come
from the same speaker reading the same Harvard sentence. For Pain vs.\ No Pain,
the model sees one No-pain clip and one Pain clip and must identify the Pain
clip. In this binary setting, each physical pair is shown in both orders. For 3-class
severity, the first recording is a reference with a known Mild label, and the model classifies the second recording. The order of the two clips is randomized with a fixed seed.
 
\begin{promptbox}{Pain vs.\ No Pain, Pairwise, Audio}{prompt:bin-pair-audio}
Listen to two recordings of the same speaker reading the same sentence and judge in which recording the speaker is more likely to be in pain.
 
Pain is self-reported by the speaker on a scale from 0 to 10. For this task, ratings 0-3 are classified as "No pain" and ratings 4-10 as "Pain".
 
One clip is No pain and the other is Pain. Compare the vocal expression between the two clips.
 
Clip 1:\newline
\aud{[AUDIO: clip\_\allowbreak{}1\_\allowbreak{}audio]}
 
Clip 2:\newline
\aud{[AUDIO: clip\_\allowbreak{}2\_\allowbreak{}audio]}
 
Based on the vocal expression, in which clip is the speaker more likely to be in pain?\newline
A. Clip 1\newline
B. Clip 2
 
Answer with a single letter (A or B) and nothing else. If uncertain, choose the more likely option.
\end{promptbox}
 
\begin{promptbox}{Pain vs.\ No Pain, Pairwise, Text+Audio}{prompt:bin-pair-ta}
Listen to two recordings of the same speaker reading the same sentence and read the transcription for each recording. Judge in which recording the speaker is more likely to be in pain.
 
Pain is self-reported by the speaker on a scale from 0 to 10. For this task, ratings 0-3 are classified as "No pain" and ratings 4-10 as "Pain".
 
One clip is No pain and the other is Pain. Compare the vocal and text expression between the two clips. The transcriptions may not fully match the recordings verbatim.
 
Clip 1:\newline
\aud{[AUDIO: clip\_\allowbreak{}1\_\allowbreak{}audio]}\newline
Transcription:\newline
\ph{\{clip\_\allowbreak{}1\_\allowbreak{}transcript\}}
 
Clip 2:\newline
\aud{[AUDIO: clip\_\allowbreak{}2\_\allowbreak{}audio]}\newline
Transcription:\newline
\ph{\{clip\_\allowbreak{}2\_\allowbreak{}transcript\}}
 
Based on the vocal and text expression, in which clip is the speaker more likely to be in pain?\newline
A. Clip 1\newline
B. Clip 2
 
Answer with a single letter (A or B) and nothing else. If uncertain, choose the more likely option.
\end{promptbox}
 
\begin{promptbox}{3-Class, Mild Reference, Audio}{prompt:tri-ref-audio}
Listen to two recordings of the same speaker reading the same sentence and judge the target's pain severity.
 
Pain is self-reported by the speaker on a scale from 0 to 10. For this task, ratings 0-3 are classified as "Mild", 4-6 as "Moderate", and 7-10 as "Severe".
 
The reference is labeled Mild. Compare the vocal expression of the target with the mild reference. Use the reference as a baseline and only classify the target.
 
Reference - known label: Mild\newline
\aud{[AUDIO: reference\_\allowbreak{}audio]}
 
Target:\newline
\aud{[AUDIO: target\_\allowbreak{}audio]}
 
Based on the vocal expression, which is more likely for the speaker in the target recording?\newline
A. Mild\newline
B. Moderate\newline
C. Severe
 
Answer with a single letter (A, B, or C) and nothing else. If uncertain, choose the more likely option.
\end{promptbox}
 
\begin{promptbox}{3-Class, Mild Reference, Text+Audio}{prompt:tri-ref-ta}
Listen to two recordings of the same speaker reading the same sentence and read the transcription for each recording. Judge the target's pain severity.
 
Pain is self-reported by the speaker on a scale from 0 to 10. For this task, ratings 0-3 are classified as "Mild", 4-6 as "Moderate", and 7-10 as "Severe".
 
The reference is labeled Mild. Compare the vocal and text expression of the target with the mild reference. Use the reference as a baseline and only classify the target. The transcriptions may not fully match the recordings verbatim.
 
Reference - known label: Mild\newline
\aud{[AUDIO: reference\_\allowbreak{}audio]}\newline
Transcription:\newline
\ph{\{reference\_\allowbreak{}transcript\}}
 
Target:\newline
\aud{[AUDIO: target\_\allowbreak{}audio]}\newline
Transcription:\newline
\ph{\{target\_\allowbreak{}transcript\}}
 
Based on the vocal and text expression, which is more likely for the speaker in the target recording and transcription?\newline
A. Mild\newline
B. Moderate\newline
C. Severe
 
Answer with a single letter (A, B, or C) and nothing else. If uncertain, choose the more likely option.
\end{promptbox}
 
\subsection{20-shot Prompts}
\label{app:prompts-20shot}
 
These prompts are used for the Neutral + 20-shot columns. The 20 labeled
examples and the target all come from the same speaker. Choices are randomly shuffled.
 
\begin{promptbox}{Pain vs.\ No Pain, 20-shot, Audio}{prompt:bin-20-audio}
Listen to the recordings and judge whether the target speaker is in pain.
 
Pain is self-reported by the speaker on a scale from 0 to 10. For this task, ratings 0-3 are classified as "No pain" and ratings 4-10 as "Pain".
 
The 20 labeled examples and the unlabeled target are from the same speaker. Use the examples to learn the speaker's vocal expression with and without pain.
 
Labeled examples:\newline
\ph{\{binary\_\allowbreak{}labeled\_\allowbreak{}audio\_\allowbreak{}examples\}}
 
Target:\newline
\aud{[AUDIO: target\_\allowbreak{}audio]}
 
Classify only the target.
 
Based on the vocal expression, which is more likely for the speaker in the target recording?\newline
A. Pain\newline
B. No pain
 
Answer with a single letter (A or B) and nothing else. If uncertain, choose the more likely option.
\end{promptbox}
 
\begin{promptbox}{Example Format for Prompt~\ref{prompt:bin-20-audio}}{prompt:bin-20-audio-ex}
Example \ph{\{i\}} - known label: \ph{\{label\}}\newline
\aud{[AUDIO: support\_\allowbreak{}audio\_\allowbreak{}\{i\}]}
\end{promptbox}
 
\begin{promptbox}{Pain vs.\ No Pain, 20-shot, Text+Audio}{prompt:bin-20-ta}
Listen to the recordings and read their transcriptions to judge whether the target speaker is in pain.
 
Pain is self-reported by the speaker on a scale from 0 to 10. For this task, ratings 0-3 are classified as "No pain" and ratings 4-10 as "Pain".
 
The 20 labeled examples and the unlabeled target are from the same speaker. Use these examples to learn the speaker's vocal and text expression with and without pain. The transcriptions may not fully match the recordings verbatim.
 
Labeled examples:\newline
\ph{\{labeled\_\allowbreak{}audio\_\allowbreak{}transcript\_\allowbreak{}examples\}}
 
Target:\newline
\aud{[AUDIO: target\_\allowbreak{}audio]}\newline
Transcription: \ph{\{target\_\allowbreak{}transcript\}}
 
Classify only the target.
 
Based on the vocal and text expression, which is more likely for the speaker in the target recording and transcription?\newline
A. Pain\newline
B. No pain
 
Answer with a single letter (A or B) and nothing else. If uncertain, choose the more likely option.
\end{promptbox}
 
\begin{promptbox}{Example Format for Prompt~\ref{prompt:bin-20-ta}}{prompt:bin-20-ta-ex}
Example \ph{\{i\}} - known label: \ph{\{label\}}\newline
\aud{[AUDIO: support\_\allowbreak{}audio\_\allowbreak{}\{i\}]}\newline
Transcription: \ph{\{support\_\allowbreak{}transcript\}}
\end{promptbox}
 
\begin{promptbox}{3-Class, 20-shot, Audio}{prompt:tri-20-audio}
Listen to the recordings and judge the target speaker's pain severity.
 
Pain is self-reported by the speaker on a scale from 0 to 10. For this task, ratings 0-3 are classified as "Mild", 4-6 as "Moderate", and 7-10 as "Severe".
 
The 20 labeled examples and the target are from the same speaker. Use the examples to learn how the speaker's vocal expression varies across these pain levels.
 
Labeled examples:\newline
\ph{\{three\_\allowbreak{}class\_\allowbreak{}labeled\_\allowbreak{}audio\_\allowbreak{}examples\}}
 
Target:\newline
\aud{[AUDIO: target\_\allowbreak{}audio]}
 
Classify only the target.
 
Based on the vocal expression, which is more likely for the speaker in the target recording?\newline
A. Mild\newline
B. Moderate\newline
C. Severe
 
Answer with a single letter (A, B, or C) and nothing else. If uncertain, choose the more likely option.
\end{promptbox}
 
\begin{promptbox}{Example Format for Prompt~\ref{prompt:tri-20-audio}}{prompt:tri-20-audio-ex}
Example \ph{\{i\}} - known label: \ph{\{label\}}\newline
\aud{[AUDIO: support\_\allowbreak{}audio\_\allowbreak{}\{i\}]}
\end{promptbox}
 
\begin{promptbox}{3-Class, 20-shot, Text+Audio}{prompt:tri-20-ta}
Listen to the recordings and read their assigned transcriptions to judge the target speaker's pain severity.
 
Pain is self-reported by the speaker on a scale from 0 to 10. For this task, ratings 0-3 are classified as "Mild", 4-6 as "Moderate", and 7-10 as "Severe".
 
The 20 labeled examples and the unlabeled target are from the same speaker. Use these examples to learn the speaker's vocal and text expression at these pain levels. The transcriptions may not fully match the recordings verbatim.
 
Labeled examples:\newline
\ph{\{three\_\allowbreak{}class\_\allowbreak{}labeled\_\allowbreak{}audio\_\allowbreak{}transcript\_\allowbreak{}examples\}}
 
Target:\newline
\aud{[AUDIO: target\_\allowbreak{}audio]}\newline
Transcription: \ph{\{target\_\allowbreak{}transcript\}}
 
Classify only the target.
 
Based on the vocal and text expression, which is more likely for the speaker in the target recording and transcription?\newline
A. Mild\newline
B. Moderate\newline
C. Severe
 
Answer with a single letter (A, B, or C) and nothing else. If uncertain, choose the more likely option.
\end{promptbox}
 
\begin{promptbox}{Example Format for Prompt~\ref{prompt:tri-20-ta}}{prompt:tri-20-ta-ex}
Example \ph{\{i\}} - known label: \ph{\{label\}}\newline
\aud{[AUDIO: support\_\allowbreak{}audio\_\allowbreak{}\{i\}]}\newline
Assigned transcription: \ph{\{support\_\allowbreak{}transcript\}}
\end{promptbox}

\subsection{Acoustic Contrast Perception Test Prompt}
\label{app:prompts-contrast}

This prompt is used for the acoustic perception test
(Figure~\ref{fig:contrast_heatmap}; Appendix~\ref{app:contrast-dataset}).
Each trial presents the low and high renderings of the same recording as two
separate audio inputs, and the model must choose the clip that is higher on the
manipulated cue. The same template is used for every model; only
\ph{\{question\}} changes with the cue (Table~\ref{tab:contrast-questions}).

\begin{promptbox}{Acoustic Contrast, Audio}{prompt:contrast}
Listen to two clips of the same speaker saying the same sentence.

Clip A:\newline
\aud{[AUDIO: clip\_\allowbreak{}a\_\allowbreak{}audio]}

Clip B:\newline
\aud{[AUDIO: clip\_\allowbreak{}b\_\allowbreak{}audio]}

\ph{\{question\}}

A. Clip A\newline
B. Clip B

Answer A or B only.
\end{promptbox}

\begin{table}[t]
\centering
\small
\begin{tabular}{@{}lp{0.6\columnwidth}@{}}
\toprule
\textbf{Cue} & \textbf{Question} \\
\midrule
Pitch level       & Which clip has the higher mean pitch? \\
Pitch variability & Which clip has more pitch variation? \\
Loudness level    & Which clip is louder overall? \\
Loudness dynamics & Which clip has greater loudness variation? \\
Spectral tilt     & Which clip sounds brighter or sharper? \\
\bottomrule
\end{tabular}
\caption{Cue-specific questions substituted into the \texttt{\{question\}} slot of
Prompt~\ref{prompt:contrast}. The same questions were shown to human listeners.}
\label{tab:contrast-questions}
\end{table}

The correct answer is always the high-magnitude clip, and its position (A or B) is
randomized across trials. Responses that cannot be parsed as A or B are counted as incorrect.

\subsection{Prompt Variants with Explicit Acoustic Cue}
\label{app:prompts-acoustic}

Every prompt above has an acoustic-measurement variant, used for the treatment setting in Table \ref{tab:acoustic-cue-delta-neutral} and Table \ref{tab:acoustic-cue-delta-zeroshot}. Each variant is obtained from the corresponding prompt with three changes:
\begin{itemize}
    \item The feature-definition block (Prompt~\ref{prompt:feat-def}) is inserted after the pain-scale description;
    \item An acoustic measurement block (Prompt~\ref{prompt:feat-val}) is inserted directly after every audio input, i.e., after the target, after each clip in pairwise prompts, after the reference, and after each labeled example in 20-shot prompts, with values extracted from that recording;
    \item ``Based on the vocal expression'' in the final question is replaced by ``Based on the vocal expression and textual acoustic measurements''. In Text+Audio prompts, this becomes ``Based on the vocal and text expression and textual acoustic measurements''.
\end{itemize}
All other wording, choices, and shuffling are unchanged. Prompt~\ref{prompt:bin-acoustic} shows the full variant of Prompt~\ref{prompt:bin-zs-audio}.

\begin{promptbox}{Feature Definition Block}{prompt:feat-def}
Feature definitions:\newline
- Pitch level: median fundamental frequency of voiced frames, in semitones relative to 27.5 Hz\newline
- Pitch variability: coefficient of variation of F0 over voiced frames, unitless\newline
- Loudness level: mean perceptual loudness over the recording, openSMILE units, uncalibrated\newline
- Loudness dynamics: mean rate at which loudness decreases during falling segments of the loudness contour\newline
- Voice quality: mean level difference in dB between the first and second harmonic of F0, voiced frames\newline
- Spectral flux: mean frame-to-frame change of the spectrum over the recording\newline
- MFCC1: mean of the first mel-frequency cepstral coefficient\newline
- Spectral tilt: mean slope of the spectrum between 0 and 500 Hz, voiced frames
\end{promptbox}

\begin{promptbox}{Measurement Block}{prompt:feat-val}
Acoustic measurements for this recording:\newline
- Pitch level: \ph{\{pitch\_\allowbreak{}median\}}\newline
- Pitch variability: \ph{\{pitch\_\allowbreak{}cv\}}\newline
- Loudness level: \ph{\{loudness\_\allowbreak{}mean\}}\newline
- Loudness dynamics: \ph{\{loudness\_\allowbreak{}fall\}}\newline
- Voice quality: \ph{\{h1h2\}}\newline
- Spectral flux: \ph{\{flux\}}\newline
- MFCC1: \ph{\{mfcc1\}}\newline
- Spectral tilt: \ph{\{slope\_\allowbreak{}0\_\allowbreak{}500\}}
\end{promptbox}

\begin{promptbox}{Pain vs.\ No Pain, Audio + Acoustic Measurements}{prompt:bin-acoustic}
Listen to the target recording and judge whether the speaker is in pain.

Pain is self-reported by the speaker on a scale from 0 to 10. For this task, ratings 0-3 are classified as "No pain" and ratings 4-10 as "Pain".

\ph{[Feature Definition Block]}

Target:\newline
\aud{[AUDIO: target\_\allowbreak{}audio]}

\ph{[Measurement Block]}

Based on the vocal expression and textual acoustic measurements, which is more likely for the speaker in the target recording?\newline
A. Pain\newline
B. No pain

Answer with a single letter (A or B) and nothing else. If uncertain, choose the more likely option.
\FloatBarrier 
\end{promptbox}

\subsection{Pairwise Prompts with Raw Acoustic Measurements for LISTEN Dataset}
\label{app:prompts-listen-pairwise-acoustic}

We created an acoustic-measurement pairwise variant on original prompt template for LISTEN dataset, used for the treatment setting in Table~\ref{tab:listen-pairwise}. The measurement block (Prompt~\ref{prompt:feat-val-listen}) reports eight standard prosodic and voice-quality descriptors from eGeMAPSv02 that are widely
used in speech emotion recognition.

Each variant is obtained from the corresponding prompt with three changes:
\begin{itemize}
    \item The feature-definition block (Prompt~\ref{prompt:feat-def-listen}) is inserted at the beginning of the user message.
    \item An acoustic measurement block (Prompt~\ref{prompt:feat-val-listen}) is inserted directly after every audio input.
    \item The original response-format instruction ``Respond with only the letter (A, B, C, D, E, F, G, H):'', is replaced by the explicit answer-format instruction.
\end{itemize}
Prompt~\ref{prompt:listen-pairwise-acoustic} shows the full acoustic-measurement variant of the pairwise audio prompt.

In Audio+Text prompts, the transcript of each recording is placed after the measurement block associated with that recording. Measurements are shown as raw values and are not standardized or interpreted. If a functional is undefined for a recording, its value is displayed as \texttt{unavailable} rather than imputed.

\begin{promptbox}{Feature Definition Block}{prompt:feat-def-listen}
Acoustic signal definitions:\newline
- Mean pitch: mean fundamental frequency of voiced frames, in semitones relative to 27.5 Hz\newline
- Pitch variability: normalized variation of fundamental frequency over voiced frames\newline
- Mean loudness: mean perceptual loudness over the recording\newline
- Loudness variability: normalized variation of perceptual loudness over time\newline
- Harmonics-to-noise ratio: average ratio of periodic harmonic energy to noise energy, in decibels\newline
- Local shimmer: average cycle-to-cycle amplitude variation, in decibels\newline
- Local jitter: average cycle-to-cycle pitch-period variation\newline
- Spectral tilt, 0--500 Hz: slope of spectral energy from 0 to 500 Hz\newline
\end{promptbox}

\begin{promptbox}{Measurement Block}{prompt:feat-val-listen}
Raw acoustic measurements for this recording:\newline
- Mean pitch: \ph{\{F0semitoneFrom27.5Hz\_\allowbreak{}sma3nz\_\allowbreak{}amean\}}\newline
- Pitch variability: \ph{\{F0semitoneFrom27.5Hz\_\allowbreak{}sma3nz\_\allowbreak{}stddevNorm\}}\newline
- Mean loudness: \ph{\{loudness\_\allowbreak{}sma3\_\allowbreak{}amean\}}\newline
- Loudness variability: \ph{\{loudness\_\allowbreak{}sma3\_\allowbreak{}stddevNorm\}}\newline
- Harmonics-to-noise ratio: \ph{\{HNRdBACF\_\allowbreak{}sma3nz\_\allowbreak{}amean\}}\newline
- Local shimmer: \ph{\{shimmerLocaldB\_\allowbreak{}sma3nz\_\allowbreak{}amean\}}\newline
- Local jitter: \ph{\{jitterLocal\_\allowbreak{}sma3nz\_\allowbreak{}amean\}}\newline
- Spectral tilt, 0--500 Hz: \ph{\{slopeUV0-500\_\allowbreak{}sma3nz\_\allowbreak{}amean\}}
\end{promptbox}

\begin{promptbox}{Pairwise Emotion Classification, Audio + Acoustic Measurements}{prompt:listen-pairwise-acoustic}
\ph{[Feature Definition Block]}

A neutral reference recording is provided first. Use it only as an acoustic reference when classifying the target.

Reference recording:\newline
\aud{[AUDIO: reference\_\allowbreak{}audio]}

\ph{[Measurement Block for reference\_\allowbreak{}audio]}

Target recording:\newline
\aud{[AUDIO: target\_\allowbreak{}audio]}

\ph{[Measurement Block for target\_\allowbreak{}audio]}

Listen to the target recording and classify the emotion expressed in the speaker's vocal delivery.

What emotional state is reflected in the speaker's voice?\newline
A. \ph{\{choice\_\allowbreak{}A\}}\newline
B. \ph{\{choice\_\allowbreak{}B\}}\newline
C. \ph{\{choice\_\allowbreak{}C\}}\newline
D. \ph{\{choice\_\allowbreak{}D\}}\newline
\ph{\{\ldots\}}\newline
\ph{\{last\_\allowbreak{}choice\_\allowbreak{}letter\}}.
\ph{\{last\_\allowbreak{}choice\}}

Answer with a single letter from the displayed options and nothing else. If uncertain, choose the more likely option.
\FloatBarrier
\end{promptbox}

\begin{promptbox}{Pairwise Emotion Classification, Audio + Text + Acoustic Measurements}{prompt:listen-pairwise-acoustic-text}
\ph{[Feature Definition Block]}

A neutral reference recording is provided first. Use it only as an acoustic reference when classifying the target.

Reference recording:\newline
\aud{[AUDIO: reference\_\allowbreak{}audio]}

\ph{[Measurement Block for reference\_\allowbreak{}audio]}

Transcription: \ph{\{reference\_\allowbreak{}transcription\}}

Target recording:\newline
\aud{[AUDIO: target\_\allowbreak{}audio]}

\ph{[Measurement Block for target\_\allowbreak{}audio]}

Transcription: \ph{\{target\_\allowbreak{}transcription\}}

Listen to the target recording and classify the emotion expressed in the speaker's vocal delivery, taking the transcription into account as specified by the original LISTEN task.

What emotional state is reflected in the speaker's voice?\newline
A. \ph{\{choice\_\allowbreak{}A\}}\newline
B. \ph{\{choice\_\allowbreak{}B\}}\newline
C. \ph{\{choice\_\allowbreak{}C\}}\newline
D. \ph{\{choice\_\allowbreak{}D\}}\newline
\ph{\{\ldots\}}\newline
\ph{\{last\_\allowbreak{}choice\_\allowbreak{}letter\}}.
\ph{\{last\_\allowbreak{}choice\}}

Answer with a single letter from the displayed options and nothing else. If uncertain, choose the more likely option.
\FloatBarrier
\end{promptbox}

\end{document}